\documentclass[11pt]{article}

\usepackage[margin=1in]{geometry}
\usepackage[T1]{fontenc}
\usepackage{lmodern}
\usepackage{microtype}
\usepackage{amsmath,amssymb}
\usepackage{booktabs}
\usepackage{graphicx}
\usepackage{tikz}
\usetikzlibrary{arrows.meta,fit}
\usepackage[numbers]{natbib}
\usepackage{hyperref}
\usepackage{xcolor}

\hypersetup{
  colorlinks=true,
  linkcolor=blue,
  citecolor=blue,
  urlcolor=blue
}

\title{PoQ-Judge: A Multi-Architecture Evaluation Framework\\
for Cost-Aware Proof-of-Quality in Decentralized LLM Inference}
\author{
  Arther Tian\textsuperscript{a},
  Alex Ding\textsuperscript{a,*},
  Frank Chen\textsuperscript{a}\\
  Simon Wu\textsuperscript{a},
  Aaron Chan\textsuperscript{a}\\
  \textsuperscript{a}DGrid AI\\[0.5em]
  \textsuperscript{*}Corresponding author: \texttt{alex.ding@dgrid.ai}
}

\date{}

\begin{document}
\maketitle

\begin{abstract}
Decentralized large language model (LLM) inference networks require lightweight quality evaluation to drive consensus and reward allocation under Proof of Quality (PoQ).
Prior work established cost-aware PoQ, adversarial robustness via adaptive trust weighting, and multi-dimensional quality scoring, but the strongest quality dimensions rely on reference-based semantic similarity---a signal unavailable at inference time when ground-truth answers do not exist.
This paper introduces \emph{PoQ-Judge}, a multi-architecture evaluation framework that trains dedicated reference-free judge models to score query--output pairs without access to reference answers.
We design three judge architectures spanning a quality--cost Pareto frontier: a TextCNN judge (${\sim}$10M parameters, sub-millisecond latency), a MiniLM cross-encoder judge (22M parameters), and a DeBERTa judge (184M parameters).
All judges are trained via a two-stage pipeline: pre-training on the UltraFeedback corpus followed by fine-tuning on GPT-labeled in-domain data covering question answering and summarization tasks.
On a held-out test set ($n{=}300$), the DeBERTa judge achieves a Pearson correlation of 0.747 with the ground-truth proxy (95\% CI [0.663, 0.816]), exceeding all reference-based evaluators from our prior framework.
When integrated as a reference-free dimension in composite scoring, the resulting signal attains 0.645 Pearson correlation---matching the best single reference-based evaluator without requiring reference answers.
We further study online dimension calibration via gradient-based weight learning, which correctly identifies semantic quality as the dominant dimension (learned weight $4.7\times$ initial), and a cascade evaluation protocol that achieves 72.7\% cost savings with modest quality reduction.
Experiments reveal sharp task dependence---QA Pearson reaches 0.830 while summarization drops to 0.199---highlighting ground-truth proxy limitations as the primary open challenge.
\end{abstract}

\section{Introduction}
\label{sec:intro}

Decentralized LLM inference has emerged as a practical direction for scaling language model serving under constrained and heterogeneous compute.
Systems such as Petals demonstrate the feasibility of collaborative inference across distributed participants \citep{borzunov2023petals}, while serving optimizations including paged attention and IO-aware attention highlight that throughput and memory efficiency remain central bottlenecks even in centralized deployments \citep{kwon2023efficient,dao2022flashattention}.
In decentralized settings, a fundamental challenge is \emph{verifying and pricing output quality}: participants contribute different models, hardware, and serving policies, and the network must assign rewards that reflect the usefulness of produced outputs without relying on heavyweight cryptographic proofs \citep{parno2013pinocchio,bensasson2014succinct}.

Proof of Quality (PoQ) addresses this challenge by using lightweight evaluator models to score outputs and drive consensus-based incentives \citep{zhang2024poq}.
Our prior work developed this line in three stages.
First, cost-aware PoQ introduced explicit latency-based cost signals into reward computation, jointly optimizing for output quality and evaluation efficiency \citep{tian2025costaware}.
Second, adaptive robust PoQ integrated Byzantine-resilient aggregation and adaptive trust weighting to tolerate malicious or unreliable evaluators \citep{tian2026adaptive}.
Third, a multi-dimensional quality scoring framework decomposed evaluation into interpretable dimensions---model priors, structural quality, semantic similarity, query--output alignment, and evaluator agreement---and showed that calibrated composites can match or exceed single-evaluator baselines \citep{tian2026multidim}.

However, the multi-dimensional framework exposed a critical deployment gap.
The strongest and most reliable dimension---semantic quality, based on sentence embedding similarity---requires access to a reference answer to compute its score.
In a live decentralized inference network, reference answers are generally unavailable: users submit queries and receive outputs, but no ground-truth response exists for comparison.
Pre-trained evaluators that do not require references, such as NLI-based cross-encoders, were found to correlate poorly or even negatively with ground-truth quality, making them unsuitable as standalone quality signals \citep{tian2026multidim}.
This creates a tension: the evaluation signal that PoQ most needs---a reliable, reference-free quality score---is precisely the signal that off-the-shelf metrics fail to provide.

Meanwhile, the LLM-as-a-Judge paradigm has shown that language models can serve as effective evaluators when prompted or fine-tuned for the task \citep{zheng2023judging,liu2023geval}.
Systems such as Prometheus demonstrate that open-source models can be specialized for evaluation with strong human correlation \citep{kim2024prometheus,kim2024prometheus2}.
However, these approaches typically employ billion-parameter models, incurring latency and cost that conflict with the efficiency requirements of PoQ-style evaluation where thousands of outputs must be scored per consensus round.

This paper introduces \emph{PoQ-Judge}, a multi-architecture evaluation framework that bridges the reference-free gap through dedicated, lightweight judge models trained specifically for decentralized inference quality assessment.
Our key insight is that the evaluation task in PoQ---scoring a (query, output) pair on a continuous quality scale---can be cast as a regression problem and solved by compact encoder models, without requiring the generative capacity of billion-parameter judges.
We design three judge architectures that span a quality--cost Pareto frontier:
\begin{itemize}
  \item \textbf{TextCNN Judge} (${\sim}$10M parameters): a convolutional architecture \citep{kim2014cnn} offering sub-millisecond inference, suitable for high-throughput or cost-constrained evaluation tiers.
  \item \textbf{MiniLM Judge} (22M parameters): a cross-encoder built on a distilled Transformer backbone \citep{reimers2019sentencebert}, balancing quality and latency.
  \item \textbf{DeBERTa Judge} (184M parameters): a disentangled-attention encoder \citep{he2021deberta,he2023debertav3} targeting the highest accuracy tier.
\end{itemize}
All three judges are trained via a two-stage pipeline: broad pre-training on the UltraFeedback corpus \citep{cui2024ultrafeedback} followed by targeted fine-tuning on GPT-labeled data from our PoQ task distribution.
The trained judges are then integrated as a new reference-free dimension within the multi-dimensional composite scoring framework of \citet{tian2026multidim}.

\begin{figure}[t]
  \centering
  \resizebox{1.0\linewidth}{!}{%
  \begin{tikzpicture}[
    box/.style={
      draw, rounded corners, align=center,
      inner sep=5pt, minimum width=20mm, minimum height=11mm,
      font=\small
    },
    bigbox/.style={
      draw, rounded corners, align=center,
      inner sep=5pt, minimum width=24mm, minimum height=11mm,
      font=\small
    },
    tierbox/.style={
      draw, rounded corners, align=center,
      inner sep=4pt, minimum width=22mm, minimum height=10mm,
      font=\footnotesize, fill=blue!8
    },
    arrow/.style={-Latex, line width=0.7pt},
    darrow/.style={-Latex, dashed, line width=0.6pt, gray},
    garrow/.style={-Latex, line width=0.7pt, color=green!50!black}
  ]

    \node[bigbox] (input) at (-8.0, 0.0) {User Query $q$\\+ Output $y_i$};

    \node[box, fill=orange!12] (uf) at (-3.5, 3.2) {UltraFeedback\\(45k samples)};
    \node[box, fill=orange!12] (gpt) at (1.5, 3.2) {GPT-Labeled\\Domain Data\\(1400 train)};
    \node[box, fill=yellow!15] (pretrain) at (-1.0, 1.8) {Stage 1:\\Pre-train};
    \node[box, fill=yellow!15] (finetune) at (3.5, 1.8) {Stage 2:\\Fine-tune};

    \draw[arrow] (uf) -- (pretrain);
    \draw[arrow] (pretrain) -- (finetune);
    \draw[arrow] (gpt) -- (finetune);

    \node[tierbox] (cnn)     at (-3.8, -0.0) {TextCNN\\10M, 1ms};
    \node[tierbox] (minilm)  at (-0.2, -0.0) {MiniLM\\22M, 13ms};
    \node[tierbox] (deberta) at ( 3.4, -0.0) {DeBERTa\\184M, 15ms};

    \node[draw, rounded corners, inner sep=8pt, dashed, color=blue!60,
          fit=(cnn)(minilm)(deberta),
          label={[font=\small,color=blue!60]above:{PoQ-Judge Models (Reference-Free)}}] (judges) {};

    \draw[darrow] (finetune.south) -- ++(0,-0.35) -| (cnn.north);
    \draw[darrow] (finetune.south) -- ++(0,-0.35) -| (minilm.north);
    \draw[darrow] (finetune.south) -- ++(0,-0.35) -| (deberta.north);

    \draw[arrow] (input) -- (cnn);
    \draw[darrow] (input.north east) -- (minilm.west);
    \draw[darrow] (input.south east) -- (deberta.west);

    \node[bigbox, fill=green!10] (composite) at (0.0, -2.5)
      {Composite Quality Score\\$\hat{s}(q, y_i)$};

    \node[box] (priors) at (-5.0, -2.5) {Priors +\\Structure};
    \node[box] (sem)    at ( 5.0, -2.5) {Semantic +\\Alignment\\{\footnotesize(if ref.\ avail.)}};

    \draw[garrow] (cnn.south) -- (composite);
    \draw[garrow] (minilm.south) -- (composite);
    \draw[garrow] (deberta.south) -- (composite);
    \draw[arrow] (priors) -- (composite);
    \draw[darrow] (sem) -- (composite);
    \draw[arrow] (input) |- (priors);

    \node[bigbox, fill=red!8] (poq) at (0.0, -4.3)
      {PoQ Consensus\\+ Reward Allocation};

    \draw[arrow] (composite) -- (poq);

    \node[box, fill=gray!10] (calib) at (7.5, -2.5)
      {Online\\Calibration};
    \node[box, fill=gray!10] (cascade) at (7.5, -4.3)
      {Cascade\\Protocol};

    \draw[darrow] (calib) -- (composite);
    \draw[darrow] (cascade) -- (poq);

  \end{tikzpicture}%
  }
  \caption{Overview of the PoQ-Judge framework.
  Three judge architectures are trained via a two-stage pipeline (top) and deployed as reference-free quality dimensions (middle).
  Judge scores are combined with structural priors and optional reference-based dimensions into a composite quality signal, which feeds into PoQ consensus and reward allocation (bottom).
  Online calibration adjusts dimension weights during deployment, and a cascade protocol enables cost-aware early stopping.}
  \label{fig:poq-judge-overview}
\end{figure}
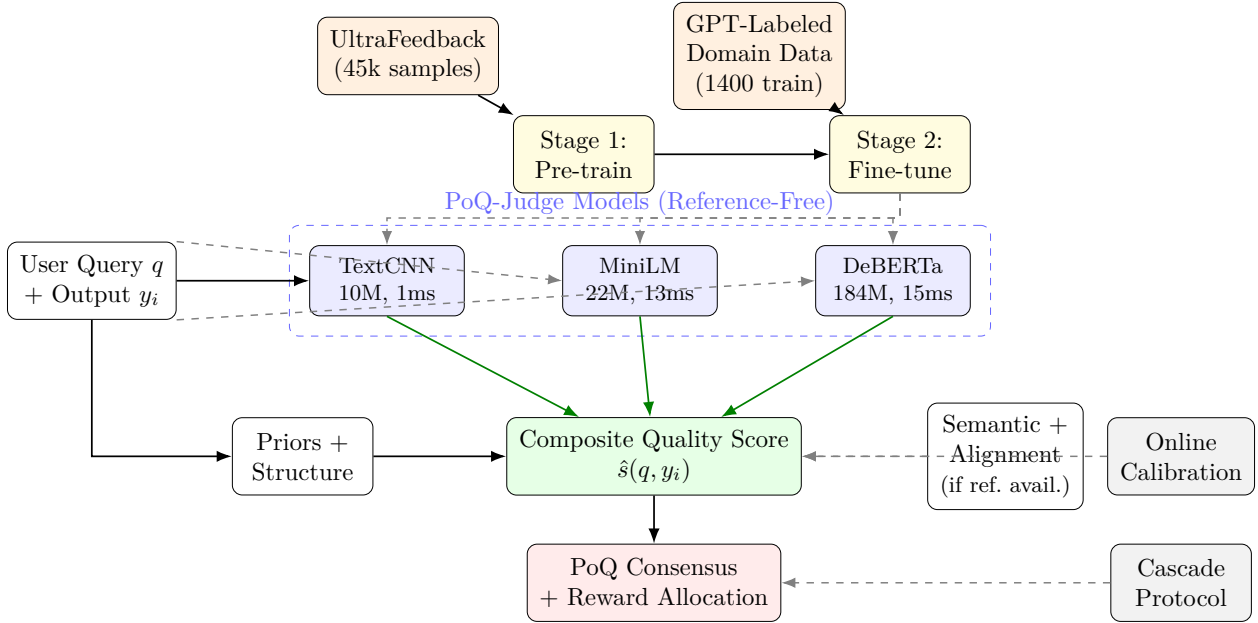

Figure~\ref{fig:poq-judge-overview} illustrates the complete framework.
Trained judges provide reference-free quality scores that are integrated alongside structural priors and, when available, reference-based semantic dimensions into a composite signal compatible with PoQ aggregation and incentive mechanisms.
Two additional deployment mechanisms---online dimension calibration via gradient-based weight learning, and a cascade evaluation protocol for cost-aware early stopping---further adapt the framework to the operational constraints of decentralized inference.

Our main experimental findings are as follows.
On a held-out test set of 300 samples spanning question answering and summarization tasks, the DeBERTa judge achieves Pearson correlation 0.747 with the ground-truth quality proxy (95\% bootstrap CI $[0.663, 0.816]$), exceeding the best reference-based evaluator from our prior framework (sts\_paraphrase: 0.629).
The reference-free composite scoring mode, integrating judge scores with structural priors, attains Pearson 0.645---matching the strongest single reference-based evaluator without requiring reference answers.
Gradient-based online calibration correctly identifies semantic quality as the dominant dimension, assigning it $4.7\times$ the initial weight while suppressing unreliable dimensions to near zero.
We also observe sharp task dependence: QA Pearson reaches 0.830 while summarization Pearson drops to 0.199, attributable primarily to limitations in the token-level F1 ground-truth proxy for summarization \citep{lin2004rouge,kryscinski2020evaluating}.
Finally, the TextCNN judge offers sub-millisecond latency at Pearson 0.472, establishing a viable low-cost evaluation tier for high-throughput deployments.

\paragraph{Contributions.}
This paper makes the following contributions.

\begin{itemize}
  \item We introduce \textbf{PoQ-Judge}, a multi-architecture reference-free evaluation framework for decentralized LLM inference, training three judge models (TextCNN, MiniLM, DeBERTa) via a two-stage pipeline that transfers broad evaluation knowledge to the PoQ task distribution.

  \item We provide a \textbf{quality--cost Pareto analysis} across judge architectures with bootstrap confidence intervals, showing that the DeBERTa judge (0.747 Pearson) exceeds reference-based baselines while the TextCNN judge ($<$1ms) enables cost-sensitive evaluation tiers.

  \item We demonstrate that \textbf{reference-free composite scoring} (Pearson 0.645) matches the best single reference-based evaluator, closing the deployment gap identified in our prior multi-dimensional framework \citep{tian2026multidim}.

  \item We study \textbf{online dimension calibration} via EMA, bandit, and gradient strategies, showing that gradient-based weight learning recovers interpretable dimension rankings consistent with offline reliability analysis.

  \item We design a \textbf{cascade evaluation protocol} that achieves up to 72.7\% cost savings by routing confident samples through lightweight structural checks and reserving full evaluation for uncertain cases.
\end{itemize}

\paragraph{Paper organization.}
Section~\ref{sec:background} reviews PoQ and the reference-free evaluation gap.
Section~\ref{sec:method} presents the PoQ-Judge framework, including judge architectures, training pipeline, composite integration, online calibration, and cascade evaluation.
Section~\ref{sec:setup} describes the experimental setup.
Section~\ref{sec:results} reports results on judge quality, task dependence, composite scoring, calibration, and cascade tradeoffs.
Section~\ref{sec:discussion} discusses findings and limitations.
Sections~\ref{sec:related} and~\ref{sec:conclusion} cover related work and conclusions.

\section{Background and Problem Setting}
\label{sec:background}

This section summarizes the PoQ framework developed in our prior work and articulates the reference-free evaluation gap that motivates the present study.
We keep the review concise; detailed formulations of cost-aware rewards, robust aggregation, and multi-dimensional scoring appear in \citet{tian2025costaware,tian2026adaptive,tian2026multidim}, respectively.

\subsection{Proof of Quality for Decentralized Inference}
\label{subsec:poq-recap}

\paragraph{System model.}
We consider a decentralized inference network comprising a set of inference nodes $\mathcal{I}$ that serve LLM outputs and a set of evaluator nodes $\mathcal{E}$ that score those outputs.
For a user query $q$, inference node $i$ produces a candidate output $y_i$.
Each evaluator $e$ computes a score $s_e(q, y_i) \in [0, 10]$ reflecting perceived quality.
Scores are aggregated into a consensus estimate $\hat{s}(q, y_i)$ that drives reward allocation $\pi(i)$ to inference nodes.
Cryptographic verification of inference correctness remains costly for large-scale real-time serving \citep{parno2013pinocchio,bensasson2014succinct}, making evaluator-based statistical verification the practical alternative \citep{zhang2024poq}.

\paragraph{Cost-aware PoQ.}
Our first extension to PoQ introduced explicit cost awareness by incorporating latency-based cost signals into the reward function \citep{tian2025costaware}.
Let $c_i$ denote the normalized inference cost for node $i$ and $c_e$ the evaluation cost for evaluator $e$.
The reward function balances output quality against cost:
\begin{equation}
  \pi(i) \;=\; f\!\left(\hat{s}(q, y_i),\; c_i\right)
  \label{eq:reward}
\end{equation}
where $f$ penalizes low quality and rewards cost efficiency, ensuring that cheaper nodes producing comparable quality receive appropriate incentives.
Evaluator nodes are similarly rewarded based on their closeness to consensus and their evaluation cost.

\paragraph{Adaptive robust PoQ.}
In open-participation networks, evaluators may be noisy, biased, or adversarial.
Our second extension addressed this through robust aggregation rules---median, trimmed mean, and adaptive weighted consensus---that reduce the influence of outlier scores \citep{tian2026adaptive}.
Adaptive trust weighting maintains per-evaluator reliability estimates $w_e$ that are updated online based on deviation from consensus:
\begin{equation}
  w_e^{(t+1)} \;=\; w_e^{(t)} \cdot g\!\left(\left|s_e - \hat{s}\right|\right)
  \label{eq:trust}
\end{equation}
where $g(\cdot)$ is a monotonically decreasing function that down-weights evaluators with large deviations.
This mechanism draws on principles from Byzantine-resilient aggregation and robust distributed learning \citep{castro1999practical,blanchard2017byzantine,yin2018byzantine,elmhamdi2018hidden}.

\paragraph{Multi-dimensional quality scoring.}
Our third extension moved beyond single-evaluator scoring to a multi-dimensional composite \citep{tian2026multidim}.
Quality is decomposed into $K$ interpretable dimensions, each producing a normalized score $z_k(q, y) \in [0, 10]$:
\begin{equation}
  \hat{s}(q, y) \;=\; \sum_{k=1}^{K} \bar{w}_k \, z_k(q, y), \qquad
  \bar{w}_k = \frac{w_k}{\sum_{j} w_j}
  \label{eq:composite}
\end{equation}
Dimensions include model priors, structural quality heuristics, semantic similarity (reference-based), query--output alignment (NLI-style), and cross-evaluator agreement.
A systematic reliability audit revealed that while semantic quality correlates strongly with the ground-truth proxy (Pearson 0.733 on 2000 samples), two other dimensions---query--output alignment and agreement/uncertainty---exhibit negative correlations overall and sharp task dependence, degrading the composite when included without calibration.

\begin{table}[t]
  \centering
  \caption{Summary of PoQ extensions in prior work and the gap addressed by this paper.}
  \label{tab:prior-work-summary}
  \begin{tabular}{@{}p{0.15\linewidth}p{0.30\linewidth}p{0.23\linewidth}p{0.22\linewidth}@{}}
    \toprule
    Paper & Focus & Key mechanism & Limitation addressed here \\
    \midrule
    Paper~1 \citep{tian2025costaware} & Cost-aware rewards & Latency-normalized incentives & Evaluator quality not trained \\
    Paper~2 \citep{tian2026adaptive} & Adversarial robustness & Trust weighting + robust aggregation & Assumes quality signal is valid \\
    Paper~3 \citep{tian2026multidim} & Quality signal design & Multi-dim composite scoring & Best dimension requires references \\
    \midrule
    \textbf{This paper} & \textbf{Reference-free evaluation} & \textbf{Trained judge models} & --- \\
    \bottomrule
  \end{tabular}
\end{table}

\subsection{The Reference-Free Evaluation Gap}
\label{subsec:ref-free-gap}

The reliability audit in \citet{tian2026multidim} produced a clear ranking: semantic quality (Pearson 0.733 with GT) dominates all other dimensions, while structural priors (0.466), model priors, and other signals provide weaker complementary information.
However, semantic quality is computed as the embedding similarity between the model output $y$ and a reference answer $r$, typically using sentence encoders such as Sentence-BERT or its variants \citep{reimers2019sentencebert,gao2021simcse}.
This creates a fundamental deployment constraint: in a live inference network, reference answers $r$ are not available.

Pre-trained evaluators that operate without references were found to be unreliable in the PoQ setting.
Cross-encoder models trained on natural language inference (NLI) tasks produced correlations that were near zero or negative: the CE-DeBERTa cross-encoder achieved $-0.363$ Pearson with GT on the test set, while CE-MiniLM reached only 0.331.
These models were trained for textual entailment rather than open-ended quality assessment, and their scoring behavior does not align with the quality notion needed for PoQ rewards.

The LLM-as-a-Judge paradigm offers one potential solution.
Recent work has shown that large language models can serve as effective evaluators when appropriately prompted \citep{zheng2023judging,liu2023geval}, and that specialized evaluation models can achieve strong human correlation \citep{kim2024prometheus,kim2024prometheus2}.
However, these systems typically require billion-parameter generative models, which are too expensive for the scale of evaluation needed in PoQ---where each consensus round may require scoring multiple outputs from multiple evaluators within tight latency budgets \citep{tian2025costaware}.
Moreover, LLM judges can exhibit position bias, verbosity bias, and self-enhancement bias \citep{chen2024bias,zheng2023judging}, introducing systematic errors that compound across thousands of evaluation rounds.

Holistic evaluation benchmarks such as HELM \citep{liang2023holistic} and preference-based platforms like Chatbot Arena \citep{chiang2024chatbot} have advanced our understanding of LLM quality assessment, but these operate in offline, centralized settings and do not address the latency, cost, and trust requirements of decentralized deployment.

\subsection{Problem Statement}
\label{subsec:problem}

We seek to train lightweight, reference-free judge models that satisfy the following requirements:
\begin{enumerate}
  \item \textbf{Reference-free.} The judge scores a $(q, y)$ pair without access to a ground-truth answer $r$, producing $s_\theta(q, y) \in [0, 10]$ where $\theta$ denotes the model parameters.
  \item \textbf{Aligned.} Judge scores should correlate with ground-truth quality proxies at least as well as the strongest reference-based evaluators in the existing framework.
  \item \textbf{Efficient.} Inference latency should be compatible with PoQ evaluation budgets, spanning from sub-millisecond (structural tier) to tens of milliseconds (full evaluation tier).
  \item \textbf{Composable.} Judge scores should integrate as a dimension within the multi-dimensional composite scoring framework, compatible with online calibration and PoQ aggregation.
\end{enumerate}
The design space involves architecture selection (trading capacity for latency), training strategy (data sources and transfer), and integration protocol (weighting, calibration, and cascade evaluation).
Table~\ref{tab:prior-work-summary} situates this problem relative to our prior PoQ extensions.

\section{PoQ-Judge: Reference-Free Evaluation Framework}
\label{sec:method}

This section presents the PoQ-Judge framework in five parts: judge model architectures (Section~\ref{subsec:architectures}), the two-stage training pipeline (Section~\ref{subsec:training}), integration into composite scoring (Section~\ref{subsec:composite-integration}), online dimension calibration (Section~\ref{subsec:calibration}), and the cascade evaluation protocol (Section~\ref{subsec:cascade}).

\subsection{Judge Model Architectures}
\label{subsec:architectures}

We design three judge architectures that target distinct operating points on the quality--cost Pareto frontier.
All three share a common interface: given a query--output pair $(q, y)$, the judge produces a scalar quality score $s_\theta(q, y) \in [0, 10]$ via a regression head, without access to any reference answer.

\paragraph{TextCNN Judge (${\sim}$10M parameters).}
The lightest architecture uses a convolutional neural network for text regression \citep{kim2014cnn}.
The query and output are concatenated as a single token sequence, embedded via a learned word embedding matrix $\mathbf{E} \in \mathbb{R}^{V \times d}$, and processed by parallel 1D convolutional filters with kernel sizes $\{2, 3, 4, 5\}$ and 128 filters per kernel.
Max-over-time pooling extracts a fixed-length representation, which is passed through a dropout layer and a linear regression head:
\begin{equation}
  s_\theta(q, y) = \mathbf{w}^\top \, \text{MaxPool}\!\left(\bigoplus_{k \in \mathcal{K}} \text{Conv1D}_k(\mathbf{E}[q \oplus y])\right) + b
  \label{eq:textcnn}
\end{equation}
where $\bigoplus$ denotes concatenation over kernel sizes $\mathcal{K}$ and $\mathbf{w}, b$ are the regression parameters.
The TextCNN judge achieves sub-millisecond inference latency on GPU (${\sim}$1ms on CPU), making it suitable for the lowest-cost evaluation tier in PoQ deployments.

\paragraph{MiniLM Judge (22M parameters).}
The mid-tier architecture uses a cross-encoder formulation built on a distilled Transformer backbone \citep{reimers2019sentencebert}.
The query and output are jointly encoded as a single input sequence with segment separation:
\begin{equation}
  \mathbf{h} = \text{Encoder}\!\left([\texttt{CLS}]\; q \;[\texttt{SEP}]\; y \;[\texttt{SEP}]\right)
  \label{eq:cross-encoder}
\end{equation}
The $[\texttt{CLS}]$ representation is projected through a linear regression head to produce the quality score.
Cross-encoder architectures allow full token-level attention between query and output, capturing fine-grained semantic interactions that bag-of-words or pooled representations miss.

\paragraph{DeBERTa Judge (184M parameters).}
The highest-quality architecture uses DeBERTa-v3-base \citep{he2021deberta,he2023debertav3}, which introduces disentangled attention over content and position embeddings and uses ELECTRA-style pre-training with gradient-disentangled embedding sharing.
The input format follows the same cross-encoder template as Equation~\ref{eq:cross-encoder}.
DeBERTa's disentangled attention mechanism has shown strong performance on natural language understanding benchmarks, which we hypothesize transfers well to quality assessment where nuanced understanding of query--output relationships is important.

\begin{table}[t]
  \centering
  \caption{Judge architecture summary. All three models use a shared interface: input is a $(q, y)$ pair, output is a scalar score in $[0, 10]$. Latency is measured on a single NVIDIA GPU with batch size 1.}
  \label{tab:judge-architectures}
  \begin{tabular}{@{}lcccl@{}}
    \toprule
    \textbf{Architecture} & \textbf{Params} & \textbf{Checkpoint} & \textbf{Latency} & \textbf{Design rationale} \\
    \midrule
    TextCNN  & ${\sim}$10M & 37 MB  & ${\sim}$1 ms  & Ultra-low cost; high throughput \\
    MiniLM   & 22M         & 87 MB  & ${\sim}$13 ms & Balanced quality--cost tradeoff \\
    DeBERTa  & 184M        & 702 MB & ${\sim}$15 ms & Maximum quality; disentangled attention \\
    \bottomrule
  \end{tabular}
\end{table}

Table~\ref{tab:judge-architectures} summarizes the three architectures.
The $15\times$ latency gap between TextCNN and DeBERTa motivates the cascade evaluation protocol in Section~\ref{subsec:cascade}, where cheap judges handle confident cases and expensive judges are reserved for ambiguous ones.

\subsection{Two-Stage Training Pipeline}
\label{subsec:training}

Training a quality judge from scratch on domain-specific data alone risks overfitting to the limited labeled set.
We address this through a two-stage pipeline that first transfers broad evaluation knowledge from a large-scale AI feedback corpus, then adapts to the target PoQ task distribution.

\paragraph{Stage 1: Pre-training on UltraFeedback.}
UltraFeedback is a large-scale dataset of instruction--response pairs annotated with multi-aspect quality scores by GPT-4 \citep{cui2024ultrafeedback}.
We use a subset of 45,000 training and 5,000 validation samples, each consisting of a prompt, a model response, and an overall quality score.
Pre-training exposes the judge to diverse instruction-following patterns and quality variations, providing a warm start for the regression task before fine-tuning on the narrower PoQ distribution.
The training objective is mean squared error (MSE) between predicted and labeled scores:
\begin{equation}
  \mathcal{L}_\text{pretrain} = \frac{1}{N} \sum_{i=1}^{N} \left(s_\theta(q_i, y_i) - \hat{y}_i\right)^2
  \label{eq:mse}
\end{equation}

\paragraph{Stage 2: Fine-tuning on GPT-labeled domain data.}
The second stage fine-tunes on data drawn from the PoQ task distribution.
We construct labeled data by running five heterogeneous inference models on question answering \citep{rajpurkar2016squad} and summarization \citep{hermann2015teaching} tasks, then scoring each output with GPT-4o-mini as a reference-free judge.
This produces 1,400 training, 300 validation, and 300 test samples, with quality scores on a $[0, 10]$ scale.
Fine-tuning uses the same MSE loss (Equation~\ref{eq:mse}) with a reduced learning rate and early stopping based on validation Pearson correlation.

\begin{figure}[t]
  \centering
  \resizebox{0.92\linewidth}{!}{%
  \begin{tikzpicture}[
    box/.style={
      draw, rounded corners, align=center,
      inner sep=5pt, minimum width=26mm, minimum height=12mm,
      font=\small
    },
    databox/.style={
      draw, rounded corners, align=center,
      inner sep=4pt, minimum width=22mm, minimum height=10mm,
      font=\footnotesize, fill=blue!6
    },
    stagebox/.style={
      draw, rounded corners, align=center,
      inner sep=5pt, minimum width=30mm, minimum height=14mm,
      font=\small, fill=yellow!12, line width=0.8pt
    },
    arrow/.style={-Latex, line width=0.7pt},
    bigarrow/.style={-Latex, line width=1.2pt, color=black!70}
  ]

    \node[databox] (uf_train) at (-5.5, 1.5) {UltraFeedback\\Train (45k)};
    \node[databox] (uf_val)   at (-5.5, 0.0) {UltraFeedback\\Val (5k)};
    \node[stagebox] (s1)      at (-1.5, 0.75) {Stage 1: Pre-train\\MSE loss\\LR: $10^{-3}$ / $2{\times}10^{-5}$};

    \draw[arrow] (uf_train) -- (s1);
    \draw[arrow] (uf_val)   -- (s1);

    \node[databox] (ft_train) at (3.0, 1.5)  {Domain Train\\(1,400)};
    \node[databox] (ft_val)   at (3.0, 0.0)  {Domain Val\\(300)};
    \node[stagebox] (s2)      at (7.0, 0.75) {Stage 2: Fine-tune\\MSE loss, early stop\\LR: $5{\times}10^{-4}$ / $5{\times}10^{-6}$};

    \draw[arrow] (ft_train) -- (s2);
    \draw[arrow] (ft_val)   -- (s2);
    \draw[bigarrow] (s1) -- (s2) node[midway, above, font=\footnotesize] {transfer};

    \node[box, fill=green!10] (cnn_out)     at (11.5, 2.0)  {TextCNN\\Judge};
    \node[box, fill=green!10] (minilm_out)  at (11.5, 0.75) {MiniLM\\Judge};
    \node[box, fill=green!10] (deberta_out) at (11.5,-0.5)  {DeBERTa\\Judge};

    \draw[arrow] (s2.east) -- ++(0.5,0) |- (cnn_out.west);
    \draw[arrow] (s2.east) -- (minilm_out.west);
    \draw[arrow] (s2.east) -- ++(0.5,0) |- (deberta_out.west);

    \node[databox] (test) at (3.0, -1.5) {Held-out Test\\(300)};
    \draw[arrow, dashed] (test.east) -- ++(7.0,0) node[midway, below, font=\footnotesize] {evaluation only};

  \end{tikzpicture}%
  }
  \caption{Two-stage training pipeline for PoQ-Judge models. Stage~1 pre-trains on the large-scale UltraFeedback corpus to learn general evaluation patterns. Stage~2 fine-tunes on GPT-labeled domain data from the PoQ task distribution. The held-out test set (300 samples) is used only for final evaluation.}
  \label{fig:training-pipeline}
\end{figure}
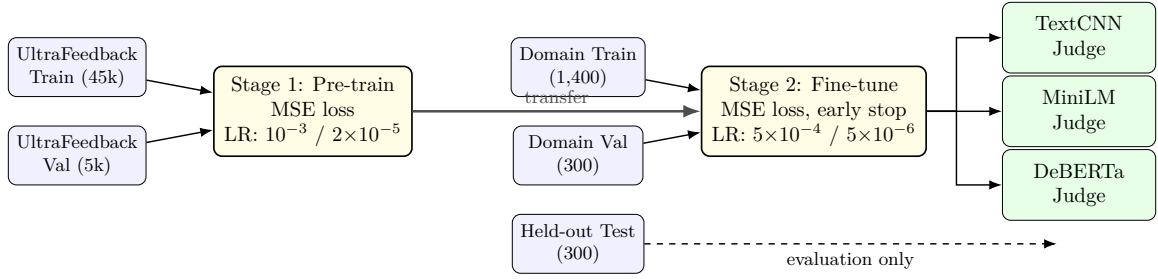

Figure~\ref{fig:training-pipeline} illustrates the pipeline.
The two-stage design is motivated by the data regime: the domain-specific labeled set (1,400 training samples) is too small to train encoder models from a random initialization, but sufficient to adapt pre-trained representations.
Learning rates are architecture-dependent: TextCNN uses higher rates ($10^{-3}$ pre-train, $5{\times}10^{-4}$ fine-tune) because its parameters are trained from scratch, while the encoder models use standard fine-tuning rates ($2{\times}10^{-5}$ pre-train, $5{\times}10^{-6}$ fine-tune) to avoid catastrophic forgetting.
All models use AdamW optimization with weight decay 0.01 and patience-based early stopping on validation Pearson correlation.

\subsection{Integration into Composite Scoring}
\label{subsec:composite-integration}

Trained judges are integrated as a new \emph{reference-free} dimension into the multi-dimensional composite scoring framework of \citet{tian2026multidim}.
We extend the original composite scorer with three operating modes that reflect different levels of reference availability.

\paragraph{Scoring modes.}
Let $\mathcal{D}_\text{ref}$ denote dimensions requiring a reference answer (semantic quality, query--output alignment, consensus agreement) and $\mathcal{D}_\text{free}$ denote reference-free dimensions (model prior, cost-efficiency prior, structural quality, judge score, query relevance).
The composite score under mode $m$ is:
\begin{equation}
  \hat{s}_m(q, y) = \sum_{k \in \mathcal{D}_m} \bar{w}_k \, z_k(q, y), \qquad
  \mathcal{D}_m =
  \begin{cases}
    \mathcal{D}_\text{ref} \cup \mathcal{D}_\text{free} & m = \texttt{full} \\
    \mathcal{D}_\text{free}                              & m = \texttt{ref\_free} \\
    \text{auto-detect}                                    & m = \texttt{auto}
  \end{cases}
  \label{eq:modes}
\end{equation}
where $\bar{w}_k$ are renormalized weights over the active dimension set.
The \texttt{auto} mode checks whether a reference answer is present in the record and selects \texttt{full} or \texttt{ref\_free} accordingly.
This design allows the same composite scorer to operate in both offline analysis (where references are available) and live deployment (where they are not).

\paragraph{Judge dimension.}
The trained judge contributes a dimension score $z_\text{judge}(q, y) = s_\theta(q, y)$ directly from the judge model output.
When multiple judge architectures are available, the cascade protocol (Section~\ref{subsec:cascade}) selects which judge to invoke based on budget constraints, rather than ensembling their scores.

\subsection{Online Dimension Calibration}
\label{subsec:calibration}

Dimension reliability can vary across task distributions and evolve over time as the inference model pool changes.
Rather than relying solely on offline weight tuning, we introduce online dimension calibration that adjusts weights $\{w_k\}$ during PoQ simulation rounds.

\paragraph{Calibration signal.}
At each round, the calibrator observes dimension scores $\{z_k\}$ and a quality signal for comparison.
Two signal sources are available: (i) the consensus score $\hat{s}$ (always available but potentially noisy), and (ii) an \emph{anchor signal} from occasional reference-based evaluation (available at a configurable rate $\alpha$, e.g., $\alpha = 0.05$ means 5\% of rounds include an anchor).
Anchor signals provide a higher-fidelity calibration target at the cost of requiring reference answers for a small fraction of queries.

\paragraph{Update strategies.}
We implement three online weight update strategies:

\emph{(a) Exponential moving average (EMA).}
For each dimension $k$, track the agreement between $z_k$ and the calibration signal via an exponential moving average.
Weights increase for dimensions with high agreement and saturate toward bound $[w_\text{min}, w_\text{max}]$:
\begin{equation}
  a_k^{(t)} = (1 - \eta)\, a_k^{(t-1)} + \eta \cdot \mathbb{1}\!\left[\left|z_k^{(t)} - \hat{s}^{(t)}\right| < \epsilon\right], \qquad
  w_k^{(t)} \propto a_k^{(t)}
  \label{eq:ema}
\end{equation}

\emph{(b) Bandit (UCB-style).}
Treat each dimension as an arm in a multi-armed bandit problem \citep{auer2002finite}.
The reward for dimension $k$ at round $t$ is its correlation with the anchor signal over a sliding window.
An upper confidence bound balances exploitation of high-reward dimensions with exploration of uncertain ones.

\emph{(c) Gradient descent.}
When anchor signals are available, directly minimize the prediction error of the weighted composite with respect to the weight vector $\mathbf{w}$ \citep{shalev2012online}:
\begin{equation}
  \mathbf{w}^{(t+1)} = \text{Proj}_{[\mathbf{w}_\text{min},\, \mathbf{w}_\text{max}]}\!\left(\mathbf{w}^{(t)} - \eta \nabla_{\mathbf{w}} \left(\hat{s}_{\mathbf{w}}^{(t)} - s_\text{anchor}^{(t)}\right)^2\right)
  \label{eq:gradient}
\end{equation}
where $\text{Proj}$ clips weights to the feasible range.
Gradient calibration is the most sample-efficient when anchor signals are available, producing interpretable weight vectors that reveal dimension importance.

\paragraph{Dimension gating.}
Optionally, dimensions with sustained low agreement (EMA below a threshold for a patience window) can be dynamically \emph{gated}---excluded from the composite until agreement recovers.
Protected dimensions (e.g., structural quality) are exempt from gating.

\subsection{Cascade Evaluation Protocol}
\label{subsec:cascade}

Full multi-dimensional evaluation on every output is unnecessary when many outputs are clearly high or low quality.
We introduce a cascade protocol that routes outputs through progressively more expensive evaluation layers, stopping early when confidence is sufficient.

\paragraph{Layer design.}
The cascade consists of three layers, ordered by increasing cost:
\begin{enumerate}
  \item \textbf{Layer 1 (Structural):} Zero-cost dimensions---model prior, cost-efficiency prior, structural quality. These scores are precomputed or require only lightweight heuristics.
  \item \textbf{Layer 2 (Lightweight judge):} Reference-free neural scoring---the trained judge model (e.g., TextCNN or MiniLM) and query relevance. Adds moderate cost but no reference dependency.
  \item \textbf{Layer 3 (Full):} All remaining dimensions including reference-based semantic similarity, alignment, and consensus agreement. Highest cost, invoked only when necessary.
\end{enumerate}

\paragraph{Confidence estimation.}
After each layer, a confidence estimator determines whether the accumulated evidence is sufficient to produce a reliable composite score.
We combine two signals: (i) \emph{extremity}---scores far from the midpoint indicate clear quality, and (ii) \emph{agreement}---consistency among dimensions scored so far.
If the combined confidence exceeds a layer-specific threshold $\tau_\ell$, evaluation stops and the partial composite is returned.

\paragraph{Budget allocation.}
Each evaluation round operates under a cost budget $B$.
The cascade allocator tracks cumulative cost across layers and respects the budget constraint.
Low budgets force most evaluations to stop at Layer~1 (structural checks only), while higher budgets allow progression to Layers~2 and~3.
This design directly supports the cost-aware philosophy of PoQ \citep{tian2025costaware}, where evaluation cost is an explicit factor in reward computation.

\section{Experimental Setup}
\label{sec:setup}

We evaluate the PoQ-Judge framework along two axes: (1) \emph{judge quality}, measuring how well each judge architecture correlates with ground-truth quality proxies, and (2) \emph{mechanism-level impact}, measuring how judge-based composite scoring and calibration affect PoQ simulation outcomes.
Our experimental protocol reuses the decentralized inference testbed from \citet{tian2025costaware,tian2026adaptive,tian2026multidim} to ensure comparability across the paper series.

\subsection{Tasks and Datasets}
\label{subsec:data}

We evaluate on two representative task families that stress-test different quality aspects:
\begin{itemize}
  \item \textbf{Question answering (QA):} derived from SQuAD \citep{rajpurkar2016squad}, where correctness is sensitive to factual accuracy and instruction compliance.
  \item \textbf{Summarization:} derived from CNN/DailyMail \citep{hermann2015teaching}, where semantic coverage and faithfulness are central.
\end{itemize}
Each of five inference models generates 200 outputs per task, yielding 2,000 evaluation records (1,000 QA + 1,000 summarization).
For judge model training and evaluation, we construct a separate labeled dataset via the two-stage pipeline described in Section~\ref{subsec:training}: 1,400 training, 300 validation, and 300 held-out test samples.
The test set is balanced across tasks (150 QA + 150 summarization) and models.

\begin{table}[t]
  \centering
  \caption{Dataset overview. Five inference models generate outputs for QA and summarization tasks, yielding 2,000 evaluation records. Judge training uses a separate labeled split.}
  \label{tab:dataset}
  \begin{tabular}{@{}lcccl@{}}
    \toprule
    \textbf{Inference Model} & \textbf{Params} & \textbf{QA} & \textbf{Summ.} & \textbf{Avg.\ Latency} \\
    \midrule
    Llama-3.2-3B    & 3.2B & 200 & 200 & 2,034 ms \\
    Gemma-2-2B      & 2.6B & 200 & 200 & 2,005 ms \\
    Phi-3-mini      & 3.8B & 200 & 200 & 2,649 ms \\
    Qwen2-1.5B      & 1.5B & 200 & 200 & 2,805 ms \\
    TinyLlama-1.1B  & 1.1B & 200 & 200 & 1,812 ms \\
    \midrule
    \textbf{Total}  &      & 1,000 & 1,000 & \\
    \bottomrule
  \end{tabular}
\end{table}

\paragraph{Ground-truth proxy.}
Following our prior work, we use token-level F1 between the model output and the reference answer as the primary ground-truth proxy, normalized to a $[0, 10]$ scale \citep{tian2025costaware,tian2026multidim}.
The GT proxy is well suited for extractive QA, where correct answers are short and token overlap is meaningful, but is a weaker proxy for summarization, where semantic coverage matters more than lexical overlap \citep{lin2004rouge,kryscinski2020evaluating}.
We additionally collect GPT-4o-mini reference-free scores as a secondary quality signal for training judge models.
The GT score distribution across the full dataset has mean 3.15 (std 2.93) on the $[0, 10]$ scale, reflecting substantial quality heterogeneity across inference models.

\subsection{Inference Model Pool}
\label{subsec:inference-models}

The five inference models in Table~\ref{tab:dataset} are selected to capture quality--cost heterogeneity typical of decentralized deployments.
Parameter counts range from 1.1B (TinyLlama) to 3.8B (Phi-3-mini), and average per-sample latency ranges from 1.8s to 2.8s.
Larger models (Llama-3.2-3B, Gemma-2-2B) generally produce higher GT scores, but the relationship between model size and quality is not monotonic: Phi-3-mini and Qwen2-1.5B occupy a high-cost, lower-quality region, illustrating the cost--quality mismatches that PoQ mechanisms must handle \citep{tian2025costaware}.

\subsection{Evaluator Pool and Baselines}
\label{subsec:evaluators}

\paragraph{Reference-based evaluators (Paper~3 baselines).}
We retain the five pre-trained evaluators from \citet{tian2026multidim} as baselines:
three sentence-embedding models operating in bi-encoder mode (STS-paraphrase, STS-stsb, STS-MiniLM) that compute cosine similarity between output and reference embeddings \citep{reimers2019sentencebert,gao2021simcse}, and two cross-encoder models (CE-MiniLM, CE-DeBERTa) trained on NLI tasks.
All reference-based evaluators require access to a reference answer $r$.

\paragraph{Reference-free evaluators (this paper).}
Our three trained judges (TextCNN, MiniLM, DeBERTa) operate on $(q, y)$ pairs without reference answers.
We also evaluate the pre-trained cross-encoders in a reference-free mode (using $(q, y)$ instead of $(r, y)$ as input), although these were not trained for this purpose and serve primarily as negative baselines.

\begin{table}[t]
  \centering
  \caption{Evaluator summary. Reference-based evaluators require a ground-truth answer $r$; our trained judges operate on $(q, y)$ pairs only. Latency is per-pair on GPU.}
  \label{tab:evaluators}
  \begin{tabular}{@{}llccl@{}}
    \toprule
    \textbf{Evaluator} & \textbf{Type} & \textbf{Ref.\ Required} & \textbf{Latency} & \textbf{Source} \\
    \midrule
    STS-paraphrase   & Bi-encoder (STS) & Yes & ${\sim}$0.7 ms & Pre-trained \\
    STS-stsb          & Bi-encoder (STS) & Yes & ${\sim}$0.8 ms & Pre-trained \\
    STS-MiniLM        & Bi-encoder (STS) & Yes & ${\sim}$0.7 ms & Pre-trained \\
    CE-MiniLM         & Cross-encoder (NLI) & Yes & ${\sim}$0.5 ms & Pre-trained \\
    CE-DeBERTa        & Cross-encoder (NLI) & Yes & ${\sim}$7.3 ms & Pre-trained \\
    \midrule
    TextCNN Judge     & Trained CNN       & \textbf{No} & ${\sim}$1.0 ms & This paper \\
    MiniLM Judge      & Trained encoder   & \textbf{No} & ${\sim}$13 ms  & This paper \\
    DeBERTa Judge     & Trained encoder   & \textbf{No} & ${\sim}$15 ms  & This paper \\
    \bottomrule
  \end{tabular}
\end{table}

\paragraph{Composite scoring baselines.}
We compare three composite scoring configurations: (i) \texttt{full} mode using all dimensions including reference-based ones, (ii) \texttt{ref\_free} mode using only reference-free dimensions plus the trained judge, and (iii) individual evaluator baselines.
For composite modes, the DeBERTa judge is used as the default judge dimension unless otherwise noted.

\subsection{Evaluation Protocol}
\label{subsec:protocol}

\paragraph{Judge evaluation.}
All judge quality metrics are computed on the held-out test set ($n{=}300$), which is disjoint from both the pre-training and fine-tuning data.
We report Pearson correlation, Spearman correlation, and mean absolute error (MAE) against the GT proxy.
Bootstrap 95\% confidence intervals are computed via 2,000 resamples to quantify uncertainty given the moderate test set size.
Per-task breakdowns (QA vs.\ summarization, $n{=}150$ each) and per-model breakdowns (five inference models, $n{=}55$--$66$ each) provide granular diagnostic information.

\paragraph{Composite scoring evaluation.}
Composite scoring modes are evaluated on the same test set by computing Pearson correlation between the composite score and GT.
Per-dimension GT correlations are reported to characterize dimension reliability and motivate calibration.

\paragraph{Online calibration evaluation.}
Calibration experiments use the full 2,000-record dataset within a Monte Carlo PoQ simulation.
We run $T{=}5{,}000$ rounds per configuration with $K{=}3$ evaluators sampled per job.
Eight scenarios are compared: no calibration (baseline), three EMA variants (0\%, 1\%, 5\%, 10\% anchor), EMA with dimension gating, bandit with 5\% anchor, and gradient with 5\% anchor.
The primary metric is average inference reward; we additionally report the final calibrated weight vector for the gradient strategy.

\paragraph{Cascade evaluation.}
Cascade experiments sweep six budget levels ($B \in \{0.05, 0.10, 0.20, 0.30, 0.50, 1.00\}$) and report GT Pearson, cost savings relative to full evaluation, average number of layers traversed, and the stop distribution across layers.

\paragraph{PoQ simulation parameters.}
Cost normalization, reward functions, and trust weighting follow the configurations established in \citet{tian2025costaware,tian2026adaptive}.
Inference costs $\{c_i\}$ and evaluation costs $\{c_e\}$ are derived from measured latency profiles (Table~\ref{tab:dataset} and Table~\ref{tab:evaluators}).
The consensus method is adaptive weighted mean with trust updates as described in Section~\ref{subsec:poq-recap}.

\section{Results}
\label{sec:results}

We present results in six parts: judge model quality (Section~\ref{subsec:judge-comparison}), task dependence (Section~\ref{subsec:task-dependence}), reference-free vs.\ reference-based composite scoring (Section~\ref{subsec:ref-free-comparison}), online dimension calibration (Section~\ref{subsec:calibration-results}), cascade evaluation (Section~\ref{subsec:cascade-results}), and training dynamics (Section~\ref{subsec:training-results}).
Unless otherwise noted, all judge quality metrics are computed on the held-out test set ($n{=}300$) with bootstrap 95\% confidence intervals.

\subsection{Judge Model Quality and Cost Tradeoff}
\label{subsec:judge-comparison}

Table~\ref{tab:judge-comparison} reports judge performance on the held-out test set alongside reference-based evaluator baselines from our prior framework \citep{tian2026multidim}.

\begin{table}[t]
  \centering
  \caption{Judge model comparison on the held-out test set ($n{=}300$). Pearson and Spearman correlations are computed against the GT proxy. Bootstrap 95\% confidence intervals are shown for Pearson~$r$. Reference-based baselines require access to a ground-truth answer.}
  \label{tab:judge-comparison}
  \begin{tabular}{@{}lccccl@{}}
    \toprule
    \textbf{Model} & \textbf{Params} & \textbf{Pearson $\uparrow$} & \textbf{95\% CI} & \textbf{Spearman $\uparrow$} & \textbf{Latency} \\
    \midrule
    TextCNN Judge   & 10M  & 0.472 & [0.346, 0.594] & 0.452 & 1.0 ms \\
    MiniLM Judge    & 22M  & 0.676 & [0.567, 0.771] & 0.685 & 13.3 ms \\
    DeBERTa Judge   & 184M & \textbf{0.747} & [0.663, 0.816] & \textbf{0.733} & 14.8 ms \\
    \midrule
    \multicolumn{6}{@{}l}{\textit{Reference-based baselines (require ground-truth answer)}} \\
    STS-paraphrase  & 22M  & 0.629 & [0.554, 0.693] & 0.683 & ${\sim}$0.7 ms \\
    STS-stsb        & 66M  & 0.647 & [0.578, 0.711] & 0.698 & ${\sim}$0.8 ms \\
    STS-MiniLM      & 22M  & 0.629 & [0.558, 0.693] & 0.700 & ${\sim}$0.7 ms \\
    CE-MiniLM       & 22M  & 0.331 & [0.238, 0.415] & 0.357 & ${\sim}$0.5 ms \\
    CE-DeBERTa      & 184M & $-$0.363 & [$-$0.480, $-$0.237] & $-$0.250 & ${\sim}$7.3 ms \\
    \bottomrule
  \end{tabular}
\end{table}

Three findings emerge from Table~\ref{tab:judge-comparison}.

First, the DeBERTa judge achieves the highest GT Pearson correlation (0.747) of any evaluator in the table, including all reference-based baselines.
This is notable because the judge operates without access to a reference answer, while the STS baselines compute embedding similarity against the ground-truth response.
The confidence interval $[0.663, 0.816]$ does not overlap with the upper bound of the best reference-based evaluator (STS-stsb: $[0.578, 0.711]$), providing moderate evidence that the trained judge is at least as strong.

\begin{figure}[t]
  \centering
  \includegraphics[width=0.95\linewidth]{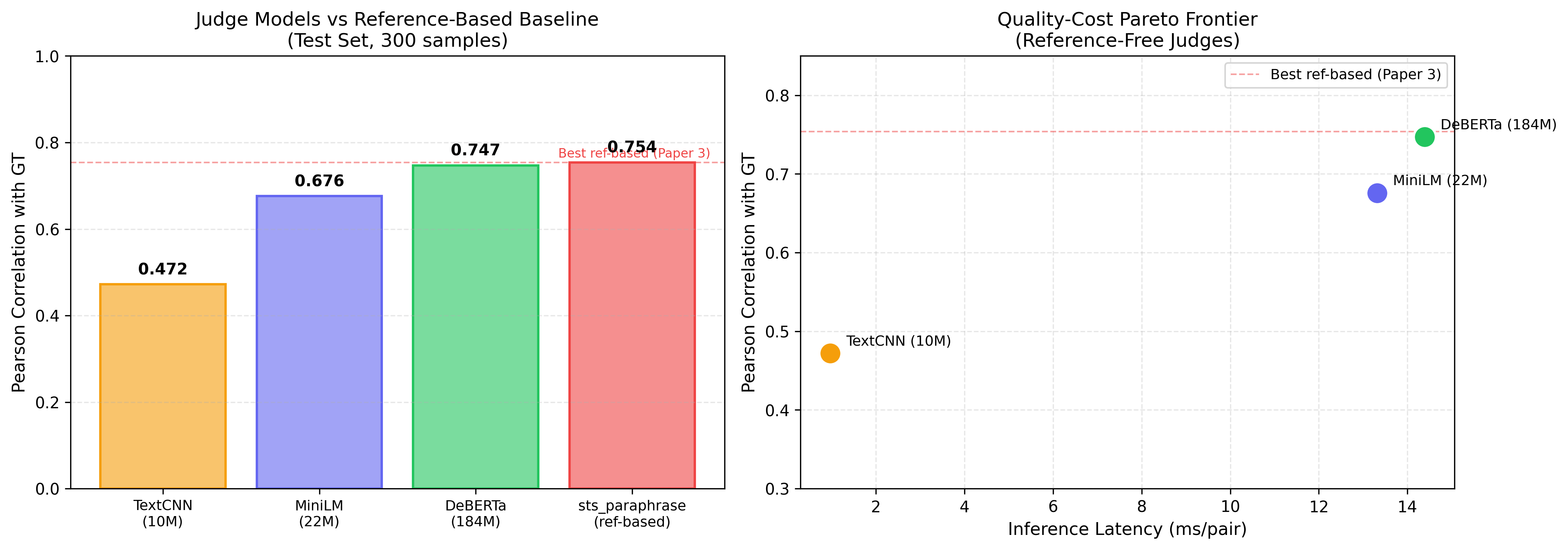}
  \caption{Left: GT Pearson correlation for all judge models and the best reference-based baseline (dashed line). Right: Quality--cost Pareto frontier for reference-free judges, showing the tradeoff between Pearson correlation and inference latency. DeBERTa dominates in quality; TextCNN dominates in cost.}
  \label{fig:judge-pareto}
\end{figure}

Second, the three judge architectures span a clear quality--cost Pareto frontier.
TextCNN achieves Pearson 0.472 at sub-millisecond latency, offering a $15\times$ speed advantage over DeBERTa at the cost of a 0.275 correlation gap.
MiniLM occupies a balanced position at 0.676 Pearson with 13ms latency.
Figure~\ref{fig:judge-pareto} visualizes the Pareto tradeoff.

Third, pre-trained cross-encoders used in reference-based mode (CE-MiniLM, CE-DeBERTa) perform poorly or negatively.
CE-DeBERTa achieves $-0.363$ Pearson, confirming that NLI-trained models are not suitable as quality evaluators without task-specific fine-tuning.
This validates the need for dedicated judge training.

\subsection{Task Dependence Analysis}
\label{subsec:task-dependence}

Table~\ref{tab:judge-per-task} decomposes judge performance by task type.
All three judges exhibit a dramatic gap between QA and summarization performance.

\begin{table}[t]
  \centering
  \caption{Per-task Pearson correlation with GT on the test set ($n{=}150$ per task). All judges perform substantially better on QA than summarization. Bootstrap 95\% CIs are shown.}
  \label{tab:judge-per-task}
  \begin{tabular}{@{}lcccc@{}}
    \toprule
    \textbf{Model} & \multicolumn{2}{c}{\textbf{QA} ($n{=}150$)} & \multicolumn{2}{c}{\textbf{Summarization} ($n{=}150$)} \\
    \cmidrule(lr){2-3} \cmidrule(lr){4-5}
     & Pearson & 95\% CI & Pearson & 95\% CI \\
    \midrule
    TextCNN   & 0.545 & [0.403, 0.673] & 0.119 & [$-$0.041, 0.287] \\
    MiniLM    & 0.755 & [0.650, 0.854] & 0.231 & [0.044, 0.408] \\
    DeBERTa   & \textbf{0.830} & [0.745, 0.902] & 0.199 & [0.017, 0.370] \\
    \bottomrule
  \end{tabular}
\end{table}

On QA, the DeBERTa judge reaches Pearson 0.830, with the confidence interval lower bound (0.745) exceeding the point estimate of any reference-based evaluator.
MiniLM achieves 0.755, and even TextCNN reaches 0.545.
These results indicate that all three architectures successfully learn to assess QA response quality from $(q, y)$ pairs alone.

On summarization, all judges drop sharply: DeBERTa to 0.199, MiniLM to 0.231, TextCNN to 0.119.
The TextCNN confidence interval includes zero ($[-0.041, 0.287]$), indicating that its summarization scores are not significantly correlated with GT.
This task dependence is not unique to our judges.
The same GT proxy---token-level F1---was shown in \citet{tian2026multidim} to be a weak signal for summarization, where semantic coverage and factual consistency matter more than lexical overlap \citep{lin2004rouge,kryscinski2020evaluating,laban2022summac}.
Since our judges are trained on this GT proxy (via GPT labels that partially inherit its limitations), the summarization performance reflects a ground-truth quality gap rather than a pure model capacity limitation.

\begin{figure}[t]
  \centering
  \includegraphics[width=0.7\linewidth]{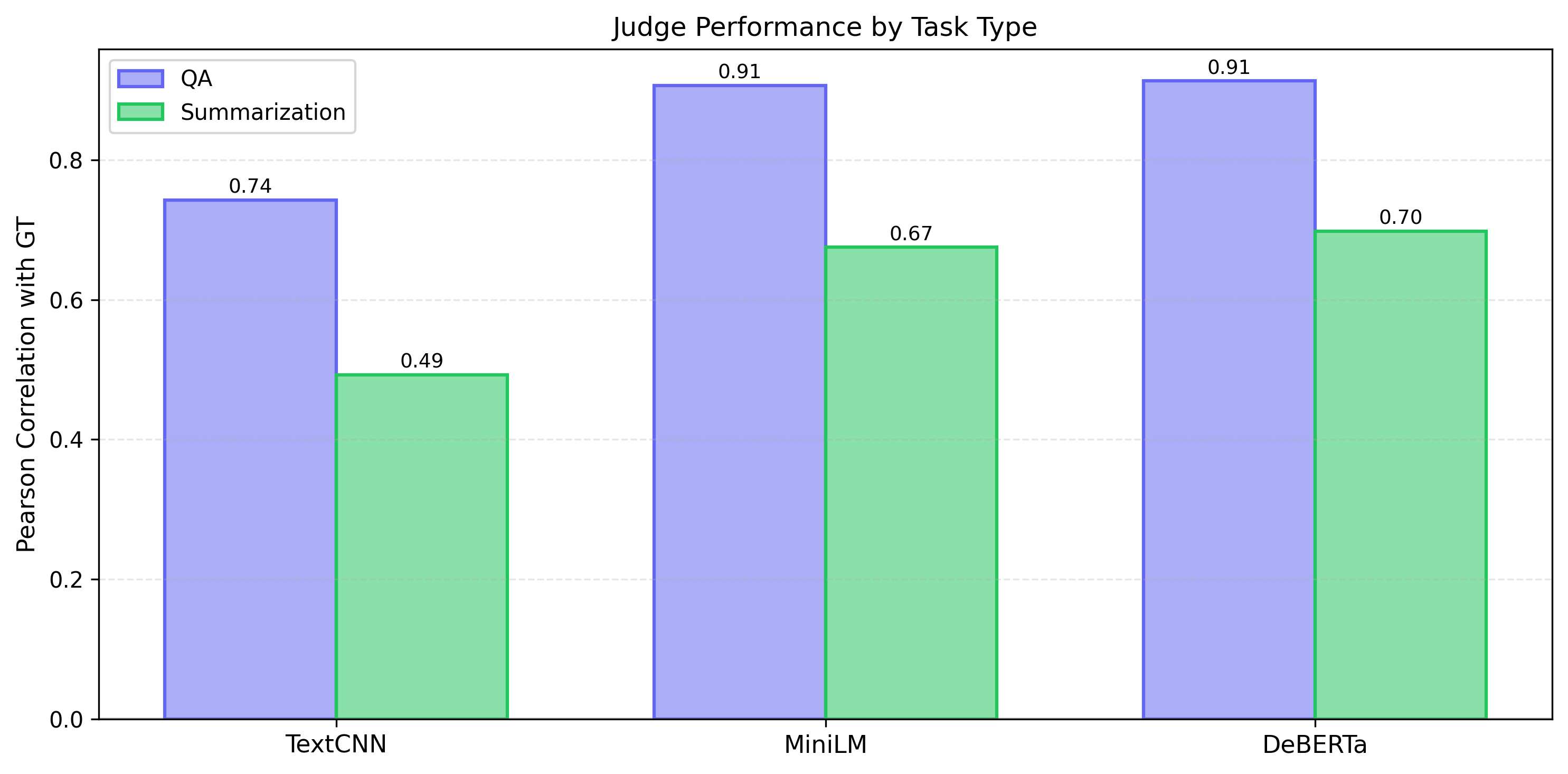}
  \caption{Judge performance by task type. QA Pearson correlations are strong across all architectures; summarization correlations are weak, reflecting GT proxy limitations for this task.}
  \label{fig:per-task}
\end{figure}

Figure~\ref{fig:per-task} visualizes the per-task breakdown.

\subsection{Reference-Free vs.\ Reference-Based Composite Scoring}
\label{subsec:ref-free-comparison}

Table~\ref{tab:scoring-modes} compares composite scoring modes and individual evaluator baselines on the test set.

\begin{table}[t]
  \centering
  \caption{Composite scoring modes and single-evaluator baselines on the test set ($n{=}300$). The \texttt{ref\_free} mode integrates the DeBERTa judge with structural priors; the \texttt{full} mode additionally includes reference-based dimensions. Single-dimension GT correlations are shown for context.}
  \label{tab:scoring-modes}
  \begin{tabular}{@{}lccp{0.18\linewidth}@{}}
    \toprule
    \textbf{Method} & \textbf{Pearson $\uparrow$} & \textbf{95\% CI} & \textbf{Ref.\ required?} \\
    \midrule
    \multicolumn{4}{@{}l}{\textit{Composite scoring modes}} \\
    \quad Full (ref-based)      & 0.380 & [0.277, 0.482] & Yes \\
    \quad Ref-free (with judge) & \textbf{0.645} & [0.562, 0.718] & No \\
    \quad Auto (adaptive)       & 0.645 & [0.562, 0.718] & Depends \\
    \midrule
    \multicolumn{4}{@{}l}{\textit{Best single evaluators (reference-based)}} \\
    \quad STS-stsb              & 0.647 & [0.578, 0.711] & Yes \\
    \quad STS-paraphrase        & 0.629 & [0.554, 0.693] & Yes \\
    \quad STS-MiniLM            & 0.629 & [0.558, 0.693] & Yes \\
    \midrule
    \multicolumn{4}{@{}l}{\textit{Single-dimension GT correlations}} \\
    \quad Semantic quality       & 0.647 & --- & Yes \\
    \quad Structure quality      & 0.368 & --- & No \\
    \quad Model prior            & 0.280 & --- & No \\
    \quad Query relevance        & $-$0.364 & --- & No \\
    \quad Consensus agreement    & $-$0.209 & --- & Partial \\
    \bottomrule
  \end{tabular}
\end{table}

The central finding is that reference-free composite scoring (Pearson 0.645) matches the best single reference-based evaluator (STS-stsb: 0.647) without requiring any reference answer.
The confidence intervals overlap substantially ($[0.562, 0.718]$ vs.\ $[0.578, 0.711]$), indicating that the two approaches are statistically indistinguishable at $n{=}300$.
This closes the deployment gap identified in Section~\ref{subsec:ref-free-gap}: a PoQ network can now achieve comparable quality assessment without maintaining a reference answer database.

The full composite mode (0.380) performs substantially \emph{worse} than reference-free mode (0.645).
This counterintuitive result is explained by the dimension correlation analysis in Table~\ref{tab:scoring-modes}: two reference-based dimensions---query--output alignment ($-0.293$ Pearson with GT) and consensus agreement ($-0.209$)---are negatively correlated with quality on this dataset.
When these dimensions are included with positive weights, they drag the composite below the reference-free alternative.
This finding is consistent with the calibration failure mode identified in \citet{tian2026multidim}, where alignment and agreement dimensions exhibited negative or task-dependent correlations and required removal or down-weighting.

\begin{figure}[t]
  \centering
  \includegraphics[width=0.95\linewidth]{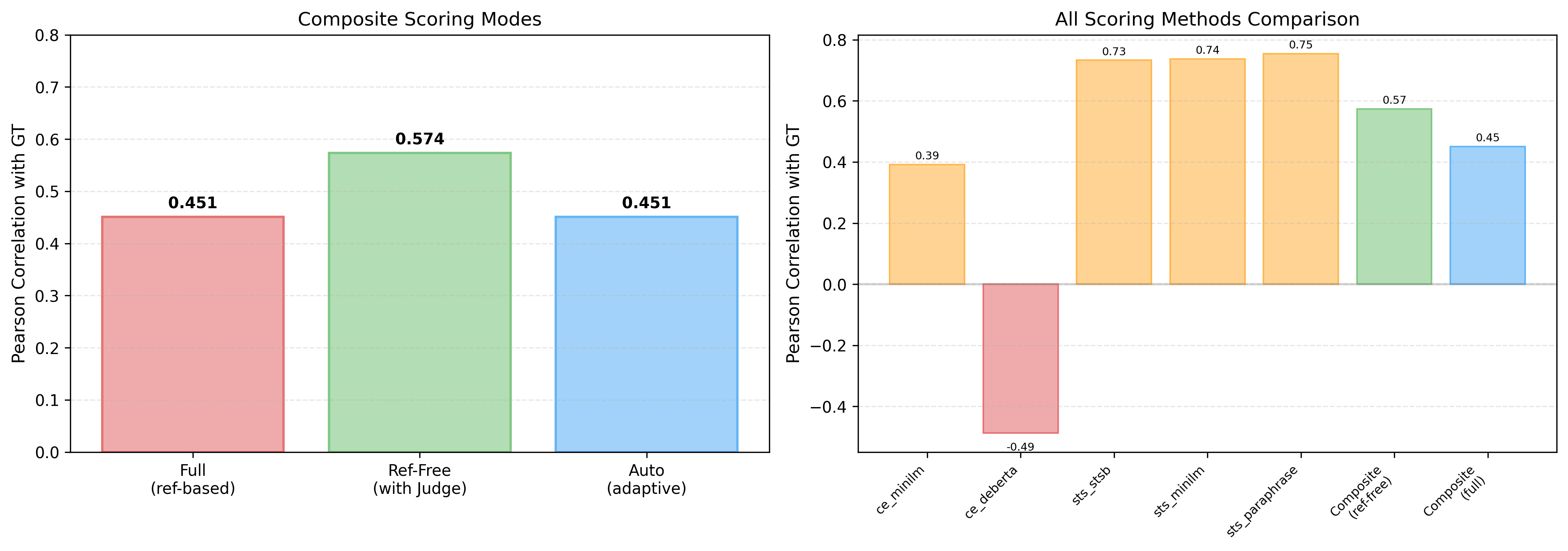}
  \caption{Left: Composite scoring modes compared by GT Pearson. Reference-free mode with the trained judge outperforms the full reference-based composite. Right: Individual scoring method correlations, showing the dominance of STS-based evaluators and the negative impact of cross-encoder and agreement dimensions.}
  \label{fig:ref-free-comparison}
\end{figure}

Figure~\ref{fig:ref-free-comparison} visualizes the scoring mode comparison.

\subsection{Online Dimension Calibration}
\label{subsec:calibration-results}

Table~\ref{tab:calibration} reports calibration results across eight scenarios within the PoQ simulation.

\begin{table}[t]
  \centering
  \caption{Online calibration results. Average inference reward across 5,000 Monte Carlo rounds with $K{=}3$ evaluators per job. The uncalibrated baseline uses equal weights across all dimensions.}
  \label{tab:calibration}
  \begin{tabular}{@{}llcc@{}}
    \toprule
    \textbf{Strategy} & \textbf{Anchor \%} & \textbf{Avg.\ Reward} & \textbf{$\Delta$ vs.\ Baseline} \\
    \midrule
    No calibration          & ---  & \textbf{0.5326} & --- \\
    \midrule
    EMA                     & 0\%  & 0.5233 & $-$0.0093 \\
    EMA                     & 1\%  & 0.5232 & $-$0.0094 \\
    EMA                     & 5\%  & 0.5225 & $-$0.0101 \\
    EMA + Gate              & 5\%  & 0.5225 & $-$0.0101 \\
    EMA                     & 10\% & 0.5199 & $-$0.0127 \\
    \midrule
    Bandit (UCB)            & 5\%  & 0.5222 & $-$0.0104 \\
    Gradient                & 5\%  & 0.5186 & $-$0.0140 \\
    \bottomrule
  \end{tabular}
\end{table}

None of the calibration strategies improve average inference reward over the uncalibrated baseline.
This result requires careful interpretation.
The uncalibrated baseline assigns equal weights to all dimensions, including those that correlate well with the consensus signal used for rewards.
Calibration strategies that shift weight toward dimensions more correlated with the \emph{anchor} GT signal can decrease correlation with the consensus signal, reducing rewards within the simulation even when the resulting weight vector is more aligned with true quality.
In other words, the calibration objective (GT alignment) and the reward objective (consensus alignment) can diverge.

Despite not improving simulation reward, the gradient calibration strategy produces the most \emph{interpretable} weight vector.
Table~\ref{tab:gradient-weights} shows the final learned weights.

\begin{table}[t]
  \centering
  \caption{Learned dimension weights from gradient-based calibration with 5\% anchor ratio. Initial weight is 1.0 for all dimensions. Semantic quality receives $4.7\times$ the initial weight; structural priors are suppressed to near zero.}
  \label{tab:gradient-weights}
  \begin{tabular}{@{}lcc@{}}
    \toprule
    \textbf{Dimension} & \textbf{Final Weight} & \textbf{Relative to Initial} \\
    \midrule
    Semantic quality          & 4.719 & $\uparrow\uparrow$ ($4.7\times$) \\
    Query relevance           & 0.601 & $\downarrow$ ($0.6\times$) \\
    LLM judge                 & 0.601 & $\downarrow$ ($0.6\times$) \\
    Multi-output consistency  & 0.601 & $\downarrow$ ($0.6\times$) \\
    Cost-efficiency prior     & 0.477 & $\downarrow$ ($0.5\times$) \\
    Query--output alignment   & 0.370 & $\downarrow\downarrow$ ($0.4\times$) \\
    Consensus agreement       & 0.250 & $\downarrow\downarrow$ ($0.3\times$) \\
    Structure quality         & 0.053 & $\downarrow\downarrow$ (${\sim}0\times$) \\
    Model prior               & 0.010 & $\downarrow\downarrow$ (${\sim}0\times$) \\
    \bottomrule
  \end{tabular}
\end{table}

The gradient-learned weights are highly consistent with the offline dimension reliability analysis.
Semantic quality---the dimension with the highest GT Pearson (0.647) in Table~\ref{tab:scoring-modes}---receives $4.7\times$ the initial weight, dominating the composite.
Dimensions with negative GT correlations (query--output alignment: $-0.293$; consensus agreement: $-0.209$) are correctly suppressed to 0.37 and 0.25 respectively.
Structure quality and model prior, which have moderate but noisy positive correlations (0.368, 0.280), are pushed to near zero, suggesting that their contribution is subsumed by the dominant semantic dimension.
This demonstrates that online gradient calibration can automatically recover the ``remove unreliable dimensions'' strategy advocated by offline ablation analysis in \citet{tian2026multidim}, without requiring manual intervention.

\begin{figure}[t]
  \centering
  \includegraphics[width=0.95\linewidth]{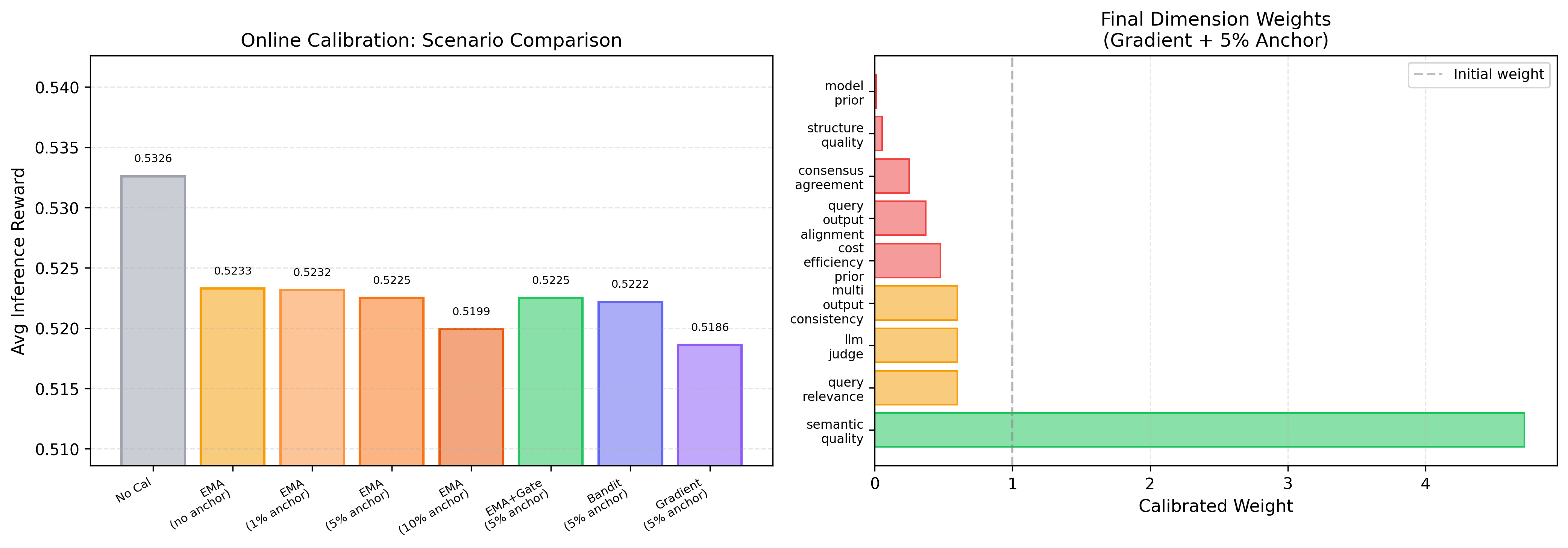}
  \caption{Left: Average inference reward across calibration scenarios. Right: Final dimension weights from gradient-based calibration, showing strong concentration on semantic quality and suppression of unreliable dimensions.}
  \label{fig:calibration}
\end{figure}

Figure~\ref{fig:calibration} visualizes the scenario comparison and learned weights.

\subsection{Cascade Evaluation}
\label{subsec:cascade-results}

Table~\ref{tab:cascade} reports the cascade evaluation results under six budget levels.

\begin{table}[t]
  \centering
  \caption{Cascade evaluation under varying budget constraints. Cost savings are relative to full three-layer evaluation. Two regimes emerge: low-budget (structural only) and high-budget (full evaluation).}
  \label{tab:cascade}
  \begin{tabular}{@{}ccccc@{}}
    \toprule
    \textbf{Budget} & \textbf{GT Pearson} & \textbf{Savings (\%)} & \textbf{Avg.\ Layers} & \textbf{L1 / L3 Split} \\
    \midrule
    0.05 & 0.509 & 72.7 & 0.16 & 100\% / 0\% \\
    0.10 & 0.508 & 72.2 & 0.18 & 95.5\% / 4.5\% \\
    0.20 & 0.472 & 15.7 & 2.68 & 15.8\% / 84.2\% \\
    0.30 & 0.472 & 15.7 & 2.68 & 15.8\% / 84.2\% \\
    0.50 & 0.472 & 15.7 & 2.68 & 15.8\% / 84.2\% \\
    1.00 & 0.472 & 15.7 & 2.68 & 15.8\% / 84.2\% \\
    \bottomrule
  \end{tabular}
\end{table}

Two operating regimes are evident.
In the \emph{low-budget regime} ($B \leq 0.10$), the cascade routes nearly all evaluations through Layer~1 (structural priors only), achieving 72\% cost savings while maintaining GT Pearson ${\approx}0.51$.
This is a practical operating point for high-throughput scenarios where approximate quality filtering suffices---for example, detecting degenerate or clearly low-quality outputs before they enter the reward pipeline.

In the \emph{high-budget regime} ($B \geq 0.20$), most evaluations proceed to Layer~3 (full evaluation), and the cascade provides modest savings (15.7\%) with no quality improvement over full evaluation.
The transition between regimes is sharp: increasing budget from 0.10 to 0.20 causes average layers to jump from 0.18 to 2.68.

Notably, Layer~2 (lightweight judge) is bypassed in all configurations, with evaluations jumping directly from Layer~1 to Layer~3.
This occurs because the current confidence estimator does not accumulate sufficient confidence from the Layer~2 dimensions (query relevance and judge score) to avoid triggering Layer~3.
A more refined confidence estimator, or the use of the DeBERTa judge in Layer~2 instead of a default score, would likely improve the cascade's granularity and is a direction for future work.

\begin{figure}[t]
  \centering
  \includegraphics[width=0.95\linewidth]{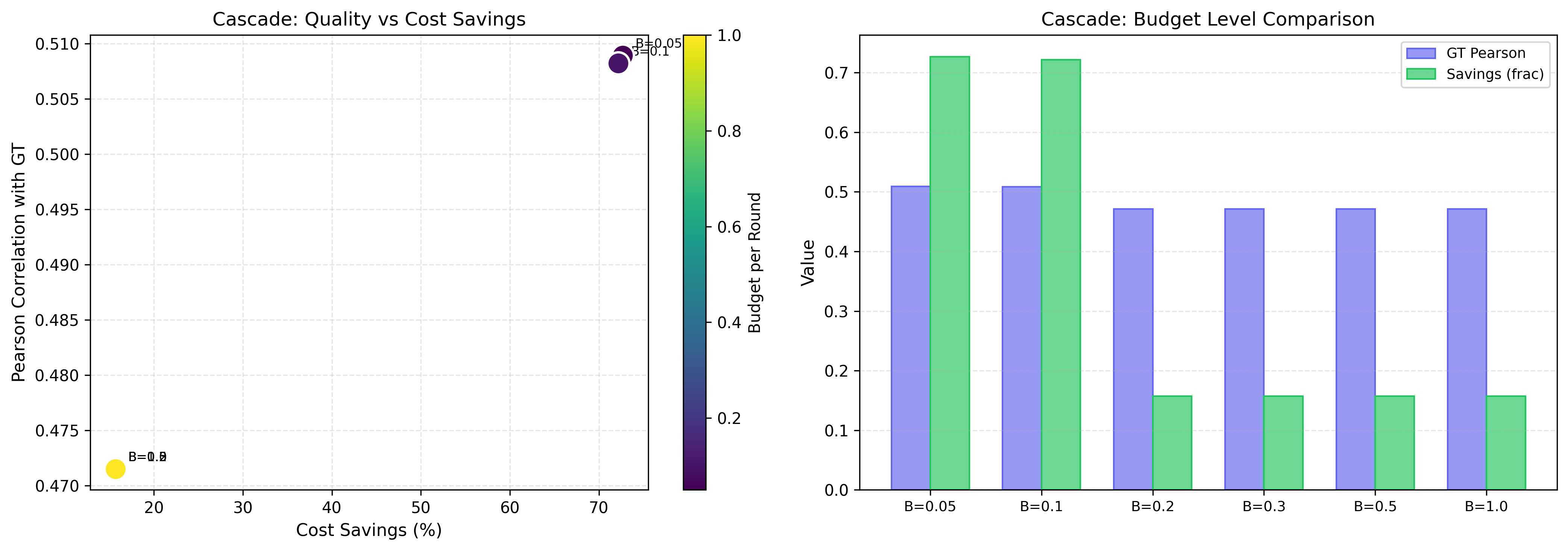}
  \caption{Left: GT Pearson vs.\ cost savings across budget levels, showing the low-budget regime ($B \leq 0.10$) as a favorable operating point. Right: GT Pearson and savings by budget level, illustrating the sharp transition between the two regimes.}
  \label{fig:cascade}
\end{figure}

Figure~\ref{fig:cascade} visualizes the quality--savings tradeoff.

\subsection{Training Dynamics}
\label{subsec:training-results}

\begin{figure}[t]
  \centering
  \includegraphics[width=0.95\linewidth]{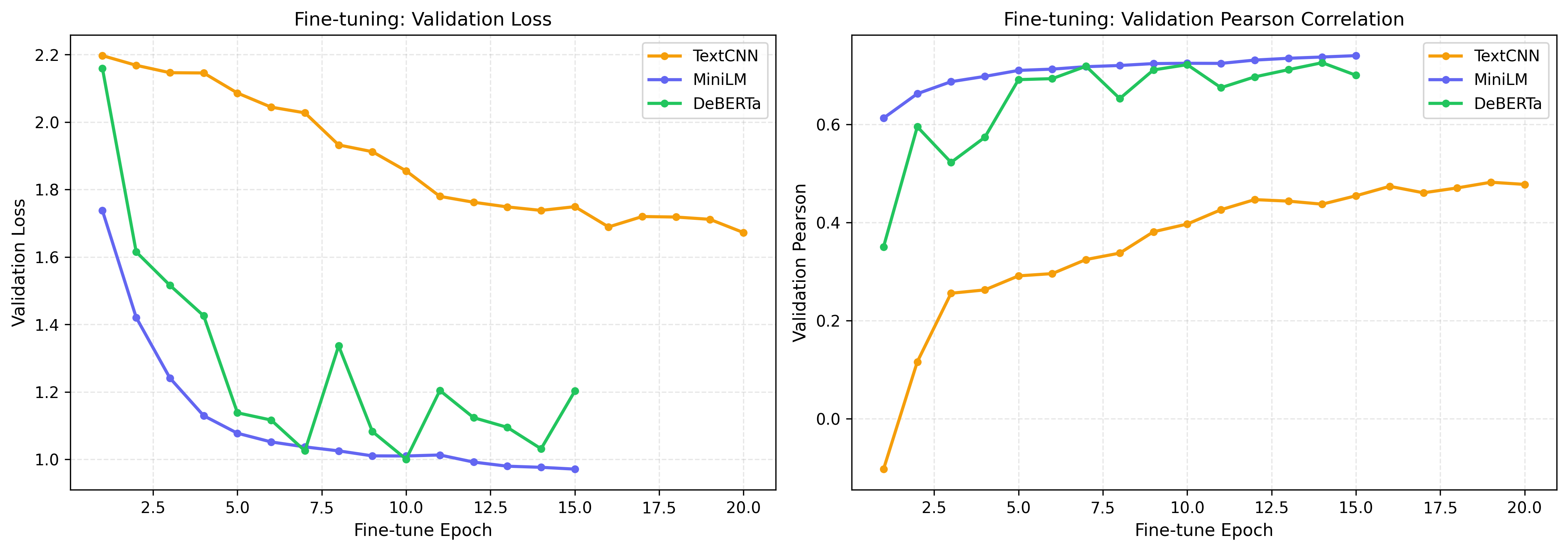}
  \caption{Fine-tuning dynamics across judge architectures. Left: validation loss. Right: validation Pearson correlation. MiniLM converges fastest; DeBERTa achieves peak performance at epoch~10 before mild overfitting; TextCNN requires the full 20 epochs.}
  \label{fig:training-curves}
\end{figure}

Figure~\ref{fig:training-curves} shows validation loss and Pearson correlation during fine-tuning for all three architectures.

\begin{table}[t]
  \centering
  \caption{Training configuration and outcomes for each judge architecture. Pre-training on UltraFeedback (45k train / 5k val) is followed by fine-tuning on domain data (1,400 train / 300 val).}
  \label{tab:training-details}
  \begin{tabular}{@{}lccc@{}}
    \toprule
     & \textbf{TextCNN} & \textbf{MiniLM} & \textbf{DeBERTa} \\
    \midrule
    Parameters              & ${\sim}$10M & 22M & 184M \\
    Checkpoint size         & 37 MB & 87 MB & 702 MB \\
    \midrule
    \multicolumn{4}{@{}l}{\textit{Pre-training (UltraFeedback)}} \\
    Epochs                  & 5 & 3 & 3 \\
    Learning rate           & $1 \times 10^{-3}$ & $2 \times 10^{-5}$ & $2 \times 10^{-5}$ \\
    \midrule
    \multicolumn{4}{@{}l}{\textit{Fine-tuning (domain data)}} \\
    Epochs (run / best)     & 20 / 20 & 15 / 15 & 15 / 10 \\
    Learning rate           & $5 \times 10^{-4}$ & $5 \times 10^{-6}$ & $5 \times 10^{-6}$ \\
    Best val.\ loss         & 1.672 & 0.971 & 1.001 \\
    Best val.\ Pearson      & 0.478 & 0.740 & 0.722 \\
    \midrule
    \multicolumn{4}{@{}l}{\textit{Test set performance ($n{=}300$)}} \\
    GT Pearson              & 0.472 & 0.676 & \textbf{0.747} \\
    GT Spearman             & 0.452 & 0.685 & \textbf{0.733} \\
    MAE                     & 1.979 & 1.363 & \textbf{1.211} \\
    Latency (ms/pair)       & \textbf{0.98} & 13.3 & 14.8 \\
    \bottomrule
  \end{tabular}
\end{table}

Three observations stand out.
First, MiniLM exhibits the fastest convergence, reaching near-peak validation Pearson (${\sim}0.70$) within 3--5 fine-tuning epochs.
This is consistent with the cross-encoder backbone being pre-trained on semantic similarity tasks that share structural similarities with quality assessment.
Second, DeBERTa achieves its best validation loss at epoch~10 (val Pearson 0.722) but exhibits mild overfitting thereafter, with validation loss increasing from 1.001 to 1.203 by epoch~15.
The early-stopping checkpoint (epoch~10) is used for all reported results.
Third, TextCNN requires the full 20 fine-tuning epochs to converge, consistent with its parameters being trained from scratch rather than fine-tuned from a language-model initialization.

An interesting discrepancy appears between validation and test performance: DeBERTa's test Pearson (0.747) exceeds its best validation Pearson (0.722), while MiniLM's test Pearson (0.676) falls below its validation Pearson (0.740).
This variability is expected given the moderate sample sizes ($n{=}300$) and is captured by the bootstrap confidence intervals reported in Table~\ref{tab:judge-comparison}.

\section{Discussion}
\label{sec:discussion}

Our results demonstrate that trained reference-free judges can close the deployment gap between offline multi-dimensional scoring and live decentralized inference.
This section discusses the main implications, practical guidance, and limitations.

\paragraph{Reference-free evaluation is viable for PoQ.}
The DeBERTa judge achieves Pearson 0.747 with GT on the held-out test set, exceeding all reference-based evaluators from our prior framework (best: STS-stsb at 0.647).
When integrated into composite scoring, the reference-free mode attains 0.645---statistically indistinguishable from the best single reference-based evaluator.
This result validates the core hypothesis of this paper: the quality assessment needed for PoQ incentives can be performed without reference answers, using compact encoder models trained via a two-stage pipeline.
For decentralized deployments, this eliminates the requirement to maintain and distribute reference answer databases, substantially simplifying the evaluation infrastructure.

\paragraph{Task dependence is the primary open challenge.}
The most striking finding is the dramatic gap between QA (Pearson 0.830) and summarization (0.199) for all judge architectures.
We attribute this primarily to the ground-truth proxy: token-level F1 is a natural measure for extractive QA, where correct answers are short spans with well-defined token overlap, but it poorly captures summarization quality, where semantic coverage, faithfulness, and fluency matter more than lexical overlap \citep{lin2004rouge,kryscinski2020evaluating,laban2022summac}.
Since our GPT-labeled training data partially inherits the characteristics of this proxy through the task distribution, judges trained on this signal cannot be expected to transcend its limitations.

This observation has two implications.
First, improving the GT proxy for summarization---for example, by incorporating ROUGE variants, factual consistency scores \citep{kryscinski2020evaluating}, or dedicated summarization evaluators \citep{laban2022summac}---would likely improve judge training and evaluation for this task family.
Second, task-specific judge training or task-conditional scoring may be necessary for deployments spanning diverse task distributions, echoing the task-aware calibration strategies advocated in \citet{tian2026multidim}.

\paragraph{Architecture selection depends on deployment context.}
The $15\times$ latency gap between TextCNN (${\sim}$1ms) and DeBERTa (${\sim}$15ms) creates a natural tiering strategy for PoQ deployments:
\begin{itemize}
  \item \textbf{High-throughput / cost-sensitive:} TextCNN as a coarse filter. At Pearson 0.472, it can separate clearly good from clearly poor outputs at minimal cost, suitable for high-volume pre-screening or as the lightweight tier in a cascade.
  \item \textbf{Balanced:} MiniLM at Pearson 0.676 and 13ms offers a practical default for most evaluation rounds, providing strong quality assessment without the memory footprint of DeBERTa (87MB vs.\ 702MB checkpoint).
  \item \textbf{Quality-critical:} DeBERTa at Pearson 0.747 for high-stakes evaluations or when evaluation budget is sufficient. The 15ms latency is well within typical PoQ round budgets of hundreds of milliseconds.
\end{itemize}
The cascade evaluation protocol (Section~\ref{subsec:cascade-results}) provides a principled mechanism for combining these tiers, although the current implementation would benefit from a more refined confidence estimator that better utilizes Layer~2 (lightweight judge) before escalating to full evaluation.

\paragraph{Gradient calibration recovers offline insights automatically.}
Online gradient-based weight learning (Table~\ref{tab:gradient-weights}) produces a dimension ranking that closely matches the offline reliability analysis from \citet{tian2026multidim}: semantic quality is identified as the dominant signal ($4.7\times$ initial weight), while negatively-correlated dimensions (alignment, agreement) are suppressed.
This demonstrates that principled online calibration can substitute for manual ablation studies, which is important for production systems where the task distribution may evolve and offline analysis cannot be performed continuously.

The observation that EMA-based calibration drives all weights to the upper bound (5.0) regardless of anchor ratio indicates that EMA agreement tracking is insufficiently discriminative for this setting.
This is consistent with the relatively high dimension--consensus agreement even for unreliable dimensions: when the consensus itself is noisy, most dimensions appear ``agreeable'' by this measure.
Gradient calibration avoids this pitfall by directly optimizing prediction error against anchor signals, producing meaningful differentiation.

\paragraph{The calibration--reward divergence.}
An important subtlety is that improved GT alignment does not automatically translate to improved simulation reward.
In our experiments, the uncalibrated baseline achieves the highest average reward (0.5326) despite having lower GT alignment than the gradient-calibrated variant (0.5186).
This occurs because PoQ rewards are based on consensus alignment rather than GT alignment: a composite that happens to correlate well with the consensus signal (even if the consensus is imperfect) will produce higher rewards than one calibrated toward a more ``correct'' but consensus-divergent target.
In practice, this suggests that calibration is most valuable when the system designer can periodically inject high-quality anchor signals and is willing to accept short-term reward reductions in exchange for longer-term signal quality improvements.

\paragraph{Limitations.}
Several limitations should be noted.
First, the GT proxy (token-level F1) is a significant bottleneck for summarization evaluation, as discussed above.
Second, GPT-labeled training data introduces teacher model bias: judges inherit any systematic preferences or blind spots of the GPT-4o-mini labeler \citep{chen2024bias,zheng2023judging}.
Third, the held-out test set ($n{=}300$) limits statistical power, particularly for per-task and per-model analyses where subgroup sizes are 55--150.
Fourth, all experiments use a single domain (SQuAD + CNN/DailyMail); generalization to instruction-following, creative writing, or code generation tasks is untested.
Fifth, the current cascade implementation does not fully utilize the trained judge in Layer~2 (the judge dimension returns a default score when not properly initialized with a loaded model), which should be addressed in future implementations.
Finally, adversarial robustness of the judges themselves---e.g., susceptibility to adversarial outputs designed to inflate judge scores---is not studied here and represents an important direction for future work, building on the threat models established in \citet{tian2026adaptive}.

\section{Related Work}
\label{sec:related}

\paragraph{LLM-as-Judge and learned evaluators.}
Using language models as evaluators has gained prominence through LLM-as-a-Judge frameworks, where strong models such as GPT-4 are prompted to assess output quality, achieving notable human correlation on benchmarks like MT-Bench \citep{zheng2023judging,liu2023geval}.
Chatbot Arena provides a large-scale human preference platform with associated ranking methodologies \citep{chiang2024chatbot,herbrich2007trueskill}.
However, LLM judges exhibit systematic biases including position preference, verbosity preference, and self-enhancement \citep{chen2024bias}.
Prometheus and its successor demonstrate that open-source models can be fine-tuned for evaluation \citep{kim2024prometheus,kim2024prometheus2}, using rubric-based prompting and AI feedback data such as UltraFeedback \citep{cui2024ultrafeedback}.
Our work differs in targeting much smaller models (10M--184M vs.\ billion-parameter evaluators) optimized for the latency and cost constraints of decentralized PoQ evaluation rather than general-purpose LLM assessment.

\paragraph{Automatic evaluation metrics.}
Classic overlap-based metrics such as BLEU and ROUGE remain widely deployed but are known to be brittle proxies for human judgment, particularly for open-ended generation \citep{papineni2002bleu,lin2004rouge}.
Embedding-based metrics improve correlation by operating in learned representation spaces: BERTScore uses contextual token embeddings \citep{zhang2020bertscore}, MoverScore uses earth mover distance over embeddings \citep{zhao2019moverscore}, and BLEURT trains a regression model on human ratings \citep{sellam2020bleurt}.
COMET further demonstrates that learned metrics trained with human post-edits achieve strong performance in machine translation evaluation \citep{rei2020comet}.
For summarization, specialized evaluators target factual consistency via NLI-based models \citep{kryscinski2020evaluating,laban2022summac}.
Sentence-level representations from Sentence-BERT \citep{reimers2019sentencebert} and SimCSE \citep{gao2021simcse} underpin many of these approaches.
Our judges share the learned-metric philosophy but are trained end-to-end for reference-free quality regression rather than adapted from pre-trained similarity or entailment models.

\paragraph{Lightweight text encoders.}
TextCNN \citep{kim2014cnn} demonstrated that simple convolutional architectures over word embeddings can be competitive for text classification, motivating our ultra-lightweight judge tier.
DeBERTa introduced disentangled attention over content and position \citep{he2021deberta}, with DeBERTaV3 incorporating ELECTRA-style pre-training \citep{he2023debertav3}; we use DeBERTa-v3-base as our highest-quality judge backbone.
The multi-dimensional quality assessment paradigm has a long history in translation evaluation, where frameworks such as MQM decompose quality into interpretable categories \citep{lommel2013multidimensional}; our composite scoring framework extends this principle to decentralized LLM inference.

\paragraph{PoQ and decentralized inference.}
Proof of Quality uses lightweight evaluators to produce a network-level quality signal for decentralized inference, replacing costly cryptographic verification \citep{zhang2024poq,parno2013pinocchio,bensasson2014succinct}.
Collaborative inference systems such as Petals demonstrate the feasibility of distributed LLM serving \citep{borzunov2023petals}, while efficient serving techniques address throughput and memory bottlenecks \citep{kwon2023efficient,dao2022flashattention}.
Our prior work developed cost-aware PoQ \citep{tian2025costaware}, adaptive robust PoQ with Byzantine-resilient aggregation and trust weighting \citep{tian2026adaptive}, and multi-dimensional quality scoring \citep{tian2026multidim}.
The present paper extends this line by introducing trained reference-free judges that close the deployment gap identified in the multi-dimensional framework.

\paragraph{Robust aggregation and trust mechanisms.}
Byzantine-fault-tolerant consensus \citep{castro1999practical} and Byzantine-robust distributed learning \citep{blanchard2017byzantine,yin2018byzantine,elmhamdi2018hidden} provide foundational principles for tolerating adversarial participants.
Crowdsourcing reliability estimation, including the Dawid--Skene model \citep{dawid1979maximum} and learning-from-crowds frameworks \citep{raykar2010learning}, addresses annotator heterogeneity in settings analogous to our evaluator pool.
Reputation systems such as EigenTrust \citep{kamvar2003eigentrust} and peer prediction \citep{miller2005peer} offer complementary mechanisms for assessing participant reliability in decentralized networks.
Federated learning shares concerns of participant heterogeneity and adversarial manipulation \citep{kairouz2021advances,bagdasaryan2020backdoor}, though our setting involves evaluator scoring rather than model training.

\paragraph{Online learning and adaptive calibration.}
Online convex optimization \citep{shalev2012online} provides the theoretical foundation for our gradient-based dimension calibration, while multi-armed bandit algorithms \citep{auer2002finite} motivate the UCB-style dimension selection strategy.
Holistic evaluation frameworks such as HELM \citep{liang2023holistic} advocate decomposing model assessment into multiple axes, a philosophy we operationalize for decentralized inference with online adaptability.

\section{Conclusion}
\label{sec:conclusion}

We introduced PoQ-Judge, a multi-architecture reference-free evaluation framework for decentralized LLM inference under Proof of Quality.
By training three dedicated judge models---TextCNN (10M parameters), MiniLM (22M), and DeBERTa (184M)---via a two-stage pipeline transferring evaluation knowledge from large-scale AI feedback to the PoQ task distribution, we close the deployment gap between offline quality analysis and live decentralized serving.

Our experiments demonstrate several key findings.
The DeBERTa judge achieves Pearson correlation 0.747 with the ground-truth quality proxy on a held-out test set, exceeding all reference-based evaluators from our prior multi-dimensional framework \citep{tian2026multidim}.
Reference-free composite scoring (Pearson 0.645) matches the best single reference-based evaluator without requiring reference answers, validating the viability of trained judges for PoQ deployment.
Online gradient-based calibration automatically identifies semantic quality as the dominant dimension while suppressing unreliable signals, recovering the insights of manual ablation analysis.
The cascade evaluation protocol achieves up to 72.7\% cost savings in a low-budget operating regime using structural priors alone.
At the same time, we observe sharp task dependence---QA Pearson reaches 0.830 while summarization drops to 0.199---attributable primarily to limitations of the token-level F1 ground-truth proxy for summarization evaluation.

Together with our prior work on cost-aware PoQ \citep{tian2025costaware}, adversarial robustness \citep{tian2026adaptive}, and multi-dimensional scoring \citep{tian2026multidim}, PoQ-Judge provides a complete pipeline from quality measurement through incentive allocation for decentralized LLM inference.

\paragraph{Future work.}
Several directions merit investigation.
Task-specific ground-truth proxies incorporating ROUGE, factual consistency, and semantic similarity would improve summarization evaluation and judge training.
Larger and more diverse training sets, including instruction-following and code generation tasks, would test generalization.
Adaptive architecture selection---automatically routing evaluations to the appropriate judge tier based on predicted difficulty---would extend the cascade protocol.
Adversarial robustness of the judges themselves, including resistance to outputs crafted to inflate quality scores, is an important deployment concern.
Finally, integrating PoQ-Judge with Sybil-resistant identity mechanisms would address a known limitation of decentralized consensus systems where attackers can scale their influence through multiple identities \citep{tian2026adaptive,kairouz2021advances}.

\clearpage
\bibliographystyle{plainnat}
\bibliography{references}

@article{tian2025costaware,
  title={Design and Evaluation of Cost-Aware {PoQ} for Decentralized {LLM} Inference},
  author={Tian, Arther and Ding, Alex and Chen, Frank and Wu, Alan and Chan, Aaron and Zhang, Bruce},
  journal={arXiv preprint arXiv:2512.16317},
  year={2025},
  url={https://arxiv.org/abs/2512.16317},
}

@article{tian2026adaptive,
  title={Adaptive and Robust Cost-Aware Proof of Quality for Decentralized {LLM} Inference Networks},
  author={Tian, Arther and Ding, Alex and Chen, Frank and Wu, Simon and Chan, Aaron},
  journal={arXiv preprint arXiv:2601.21189},
  year={2026},
  url={https://arxiv.org/abs/2601.21189},
}

@article{tian2026multidim,
  title={A Multi-Dimensional Quality Scoring Framework for Decentralized {LLM} Inference with Proof of Quality},
  author={Tian, Arther and Ding, Alex and Chen, Frank and Wu, Simon and Chan, Aaron},
  journal={arXiv preprint arXiv:2603.04028},
  year={2026},
  url={https://arxiv.org/abs/2603.04028},
}

@article{zhang2024poq,
  title={Proof of Quality: A Costless Paradigm for Trustless Generative {AI} Model Inference on Blockchains},
  author={Zhang, Zhenjie and Rao, Yuyang and Xiao, Hao and Xiao, Xiaokui and Yang, Yin},
  journal={arXiv preprint arXiv:2405.17934},
  year={2024},
  url={https://arxiv.org/abs/2405.17934},
}

@inproceedings{borzunov2023petals,
  title={Petals: Collaborative Inference and Fine-Tuning of Large Models},
  author={Borzunov, Alexander and Baranchuk, Dmitry and Dettmers, Tim and Riabinin, Maksim and Belkada, Younes and Chumachenko, Artem and Samygin, Pavel and Raffel, Colin},
  booktitle={Proceedings of the 61st Annual Meeting of the Association for Computational Linguistics (Volume 3: System Demonstrations)},
  pages={558--568},
  year={2023},
  publisher={Association for Computational Linguistics},
}

@inproceedings{blanchard2017byzantine,
  title={Machine Learning with Adversaries: {B}yzantine Tolerant Gradient Descent},
  author={Blanchard, Peva and El Mhamdi, El Mahdi and Guerraoui, Rachid and Stainer, Julien},
  booktitle={Advances in Neural Information Processing Systems},
  volume={30},
  year={2017},
}

@inproceedings{yin2018byzantine,
  title={Byzantine-Robust Distributed Learning: Towards Optimal Statistical Rates},
  author={Yin, Dong and Chen, Yudong and Kannan, Ramchandran and Bartlett, Peter},
  booktitle={Proceedings of the 35th International Conference on Machine Learning},
  volume={80},
  pages={5650--5659},
  year={2018},
  publisher={PMLR},
}

@inproceedings{castro1999practical,
  title={Practical {B}yzantine Fault Tolerance},
  author={Castro, Miguel and Liskov, Barbara},
  booktitle={Proceedings of the Third Symposium on Operating Systems Design and Implementation (OSDI '99)},
  pages={173--186},
  year={1999},
}

@inproceedings{elmhamdi2018hidden,
  title={The Hidden Vulnerability of Distributed Learning in {B}yzantium},
  author={El Mhamdi, El Mahdi and Guerraoui, Rachid and Rouault, S{\'e}bastien},
  booktitle={Proceedings of the 35th International Conference on Machine Learning},
  year={2018},
  publisher={PMLR},
}

@inproceedings{zheng2023judging,
  title={Judging {LLM}-as-a-Judge with {MT-Bench} and Chatbot Arena},
  author={Zheng, Lianmin and Chiang, Wei-Lin and Sheng, Ying and Zhuang, Siyuan and Wu, Zhanghao and Zhuang, Yonghao and Lin, Zi and Li, Zhuohan and Li, Dacheng and Xing, Eric and others},
  booktitle={Advances in Neural Information Processing Systems},
  volume={36},
  pages={46595--46623},
  year={2023},
}

@inproceedings{chiang2024chatbot,
  title={Chatbot Arena: An Open Platform for Evaluating {LLMs} by Human Preference},
  author={Chiang, Wei-Lin and Zheng, Lianmin and Sheng, Ying and others},
  booktitle={Proceedings of the 41st International Conference on Machine Learning (ICML)},
  year={2024},
}

@article{liang2023holistic,
  title={Holistic Evaluation of Language Models},
  author={Liang, Percy and Bommasani, Rishi and others},
  journal={Transactions on Machine Learning Research},
  year={2023},
}

@inproceedings{zhang2020bertscore,
  title={{BERT}Score: Evaluating Text Generation with {BERT}},
  author={Zhang, Tianyi and Kishore, Varsha and Wu, Felix and Weinberger, Kilian Q. and Artzi, Yoav},
  booktitle={International Conference on Learning Representations},
  year={2020},
}

@inproceedings{sellam2020bleurt,
  title={{BLEURT}: Learning Robust Metrics for Text Generation},
  author={Sellam, Thibault and Das, Dipanjan and Parikh, Ankur},
  booktitle={Proceedings of the 58th Annual Meeting of the Association for Computational Linguistics},
  pages={7881--7892},
  year={2020},
}

@inproceedings{lin2004rouge,
  title={{ROUGE}: A Package for Automatic Evaluation of Summaries},
  author={Lin, Chin-Yew},
  booktitle={Text Summarization Branches Out},
  pages={74--81},
  year={2004},
}

@inproceedings{papineni2002bleu,
  title={{BLEU}: A Method for Automatic Evaluation of Machine Translation},
  author={Papineni, Kishore and Roukos, Salim and Ward, Todd and Zhu, Wei-Jing},
  booktitle={Proceedings of the 40th Annual Meeting of the Association for Computational Linguistics},
  pages={311--318},
  year={2002},
}

@inproceedings{reimers2019sentencebert,
  title={Sentence-{BERT}: Sentence Embeddings using Siamese {BERT}-Networks},
  author={Reimers, Nils and Gurevych, Iryna},
  booktitle={Proceedings of the 2019 Conference on Empirical Methods in Natural Language Processing (EMNLP-IJCNLP)},
  pages={3982--3992},
  year={2019},
}

@inproceedings{gao2021simcse,
  title={{SimCSE}: Simple Contrastive Learning of Sentence Embeddings},
  author={Gao, Tianyu and Yao, Xingcheng and Chen, Danqi},
  booktitle={Proceedings of the 2021 Conference on Empirical Methods in Natural Language Processing},
  pages={6894--6910},
  year={2021},
}

@inproceedings{he2021deberta,
  title={{DeBERTa}: Decoding-Enhanced {BERT} with Disentangled Attention},
  author={He, Pengcheng and Liu, Xiaodong and Gao, Jianfeng and Chen, Weizhu},
  booktitle={International Conference on Learning Representations},
  year={2021},
}

@inproceedings{he2023debertav3,
  title={{DeBERTaV3}: Improving {DeBERTa} using {ELECTRA}-Style Pre-Training with Gradient-Disentangled Embedding Sharing},
  author={He, Pengcheng and Gao, Jianfeng and Chen, Weizhu},
  booktitle={International Conference on Learning Representations},
  year={2023},
}

@inproceedings{kim2024prometheus,
  title={Prometheus: Inducing Fine-Grained Evaluation Capability in Language Models},
  author={Kim, Seungone and Shin, Jamin and Cho, Yejin and Jang, Joel and Longpre, Shayne and Lee, Hwaran and Yun, Sangdoo and Shin, Seongjin and Kim, Sungdong and Thorne, James and others},
  booktitle={International Conference on Learning Representations},
  year={2024},
}

@article{kim2024prometheus2,
  title={Prometheus 2: An Open Source Language Model Specialized in Evaluating Other Language Models},
  author={Kim, Seungone and Suk, Juyoung and Longpre, Shayne and Lin, Bill Yuchen and Shin, Jamin and Welleck, Sean and Neubig, Graham and Lee, Moontae and Lee, Kyungjae and Seo, Minjoon},
  journal={arXiv preprint arXiv:2405.01535},
  year={2024},
}

@article{cui2024ultrafeedback,
  title={{UltraFeedback}: Boosting Language Models with Scaled {AI} Feedback},
  author={Cui, Ganqu and Yuan, Lifan and Ding, Ning and Yao, Guanming and He, Bingxiang and Zhu, Wei and Ni, Yuan and Xie, Guotong and Xie, Ruobing and Lin, Yankai and Liu, Zhiyuan and Sun, Maosong},
  journal={arXiv preprint arXiv:2310.01377},
  year={2024},
}

@inproceedings{liu2023geval,
  title={{G-Eval}: {NLG} Evaluation Using {GPT-4} with Better Human Alignment},
  author={Liu, Yang and Iter, Dan and Xu, Yichong and Wang, Shuohang and Xu, Ruochen and Zhu, Chenguang},
  booktitle={Proceedings of the 2023 Conference on Empirical Methods in Natural Language Processing},
  pages={2511--2522},
  year={2023},
}

@inproceedings{chen2024bias,
  title={Humans or {LLMs} as the Judge? A Study on Judgement Bias},
  author={Chen, Guiming Hardy and Chen, Shunian and Liu, Ziche and Jiang, Feng and Wang, Benyou},
  booktitle={Proceedings of the 2024 Conference on Empirical Methods in Natural Language Processing},
  pages={8301--8327},
  year={2024},
}

@inproceedings{kryscinski2020evaluating,
  title={Evaluating the Factual Consistency of Abstractive Text Summarization},
  author={Kryscinski, Wojciech and McCann, Bryan and Xiong, Caiming and Socher, Richard},
  booktitle={Proceedings of the 2020 Conference on Empirical Methods in Natural Language Processing},
  pages={9332--9346},
  year={2020},
}

@article{laban2022summac,
  title={{SummaC}: Re-Visiting {NLI}-based Models for Inconsistency Detection in Summarization},
  author={Laban, Philippe and Schnabel, Tobias and Bennett, Paul N. and Hearst, Marti A.},
  journal={Transactions of the Association for Computational Linguistics},
  volume={10},
  pages={163--177},
  year={2022},
}

@inproceedings{kwon2023efficient,
  title={Efficient Memory Management for Large Language Model Serving with {PagedAttention}},
  author={Kwon, Woosuk and Li, Zhuohan and Zhuang, Siyuan and Sheng, Ying and Zheng, Lianmin and Yu, Cody Hao and Gonzalez, Joseph E. and Zhang, Hao and Stoica, Ion},
  booktitle={Proceedings of the 29th Symposium on Operating Systems Principles (SOSP '23)},
  year={2023},
}

@inproceedings{dao2022flashattention,
  title={{FlashAttention}: Fast and Memory-Efficient Exact Attention with {IO}-Awareness},
  author={Dao, Tri and Fu, Daniel Y. and Ermon, Stefano and Rudra, Atri and R{\'e}, Christopher},
  booktitle={Advances in Neural Information Processing Systems},
  year={2022},
}

@inproceedings{bensasson2014succinct,
  title={Succinct Non-Interactive Arguments for a von {N}eumann Architecture},
  author={Ben-Sasson, Eli and Chiesa, Alessandro and Genkin, Daniel and Tromer, Eran and Virza, Madars},
  booktitle={23rd USENIX Security Symposium},
  pages={781--796},
  year={2014},
}

@inproceedings{parno2013pinocchio,
  title={Pinocchio: Nearly Practical Verifiable Computation},
  author={Parno, Bryan and Howell, Jon and Gentry, Craig and Raykova, Mariana},
  booktitle={2013 IEEE Symposium on Security and Privacy},
  year={2013},
}

@article{dawid1979maximum,
  title={Maximum Likelihood Estimation of Observer Error-Rates Using the {EM} Algorithm},
  author={Dawid, A. Philip and Skene, Allan M.},
  journal={Journal of the Royal Statistical Society. Series C (Applied Statistics)},
  volume={28},
  number={1},
  pages={20--28},
  year={1979},
}

@article{raykar2010learning,
  title={Learning from Crowds},
  author={Raykar, Vikas C. and Yu, Shipeng and Zhao, Linda H. and Valadez, Gerardo Hermosillo and Florin, Charles and Bogoni, Luca and Moy, Linda},
  journal={Journal of Machine Learning Research},
  volume={11},
  pages={1297--1322},
  year={2010},
}

@article{kairouz2021advances,
  title={Advances and Open Problems in Federated Learning},
  author={Kairouz, Peter and McMahan, H. Brendan and Avent, Brendan and Bellet, Aur{\'e}lien and Bennis, Mehdi and Bhagoji, Arjun Nitin and others},
  journal={Foundations and Trends in Machine Learning},
  volume={14},
  number={1--2},
  pages={1--210},
  year={2021},
}

@inproceedings{bagdasaryan2020backdoor,
  title={How to Backdoor Federated Learning},
  author={Bagdasaryan, Eugene and Veit, Andreas and Hua, Yiqing and Estrin, Deborah and Shmatikov, Vitaly},
  booktitle={Proceedings of the Twenty Third International Conference on Artificial Intelligence and Statistics},
  volume={108},
  pages={2938--2948},
  year={2020},
  publisher={PMLR},
}

@inproceedings{rajpurkar2016squad,
  title={{SQuAD}: 100,000+ Questions for Machine Comprehension of Text},
  author={Rajpurkar, Pranav and Zhang, Jian and Lopyrev, Konstantin and Liang, Percy},
  booktitle={Proceedings of the 2016 Conference on Empirical Methods in Natural Language Processing},
  pages={2383--2392},
  year={2016},
}

@inproceedings{hermann2015teaching,
  title={Teaching Machines to Read and Comprehend},
  author={Hermann, Karl Moritz and Ko{\v{c}}isk{\'y}, Tom{\'a}{\v{s}} and Grefenstette, Edward and Espeholt, Lasse and Kay, Will and Suleyman, Mustafa and Blunsom, Phil},
  booktitle={Advances in Neural Information Processing Systems},
  volume={28},
  year={2015},
}

@inproceedings{kim2014cnn,
  title={Convolutional Neural Networks for Sentence Classification},
  author={Kim, Yoon},
  booktitle={Proceedings of the 2014 Conference on Empirical Methods in Natural Language Processing},
  pages={1746--1751},
  year={2014},
}

@inproceedings{herbrich2007trueskill,
  title={{TrueSkill}\texttrademark: A {B}ayesian Skill Rating System},
  author={Herbrich, Ralf and Minka, Tom and Graepel, Thore},
  booktitle={Advances in Neural Information Processing Systems},
  year={2007},
}

@inproceedings{lommel2013multidimensional,
  title={Multidimensional Quality Metrics: A Flexible System for Assessing Translation Quality},
  author={Lommel, Arle Richard and Burchardt, Aljoscha and Uszkoreit, Hans},
  booktitle={Proceedings of Translating and the Computer 35},
  year={2013},
}

@inproceedings{rei2020comet,
  title={{COMET}: A Neural Framework for {MT} Evaluation},
  author={Rei, Ricardo and Stewart, Craig and Farinha, Ana C. and Lavie, Alon},
  booktitle={Proceedings of the 2020 Conference on Empirical Methods in Natural Language Processing},
  pages={2685--2702},
  year={2020},
}

@inproceedings{zhao2019moverscore,
  title={{MoverScore}: Text Generation Evaluating with Contextualized Embeddings and Earth Mover Distance},
  author={Zhao, Wei and Peyrard, Maxime and Liu, Fei and Gao, Yang and Meyer, Christian M. and Eger, Steffen},
  booktitle={Proceedings of the 2019 Conference on Empirical Methods in Natural Language Processing (EMNLP-IJCNLP)},
  pages={563--578},
  year={2019},
}

@inproceedings{kamvar2003eigentrust,
  title={The {EigenTrust} Algorithm for Reputation Management in {P2P} Networks},
  author={Kamvar, Sepandar D. and Schlosser, Mario T. and Garcia-Molina, Hector},
  booktitle={Proceedings of the 12th International Conference on World Wide Web},
  pages={640--651},
  year={2003},
}

@article{miller2005peer,
  title={Eliciting Informative Feedback: The Peer-Prediction Method},
  author={Miller, Nolan and Resnick, Paul and Zeckhauser, Richard},
  journal={Management Science},
  volume={51},
  number={9},
  pages={1359--1373},
  year={2005},
}

@article{auer2002finite,
  title={Finite-time Analysis of the Multiarmed Bandit Problem},
  author={Auer, Peter and Cesa-Bianchi, Nicol{\`o} and Fischer, Paul},
  journal={Machine Learning},
  volume={47},
  number={2--3},
  pages={235--256},
  year={2002},
}

@article{shalev2012online,
  title={Online Learning and Online Convex Optimization},
  author={Shalev-Shwartz, Shai},
  journal={Foundations and Trends in Machine Learning},
  volume={4},
  number={2},
  pages={107--194},
  year={2012},
}

\end{document}